\documentclass[10pt,twocolumn,letterpaper]{article}

\usepackage{iccv}
\usepackage{times}
\usepackage[pagebackref=true,breaklinks=true,letterpaper=true,colorlinks,bookmarks=false]{hyperref}
\usepackage{url}

\usepackage{placeins}
\usepackage{subfigure}
\usepackage{epsfig}
\usepackage{graphicx}
\usepackage{amsmath}
\usepackage{amssymb}
\usepackage{bbm}
\usepackage{epstopdf}
\usepackage{enumitem}
\usepackage{calc}
\usepackage{multirow}

\usepackage{algorithm}
\usepackage{algpseudocode}

\iccvfinalcopy 
\def\iccvPaperID{2281} 

\newcommand{\eat}[1]{}

\newcommand{\bm}[1]{\boldsymbol{#1}}
\newcommand{\argmax}{\ensuremath{\mathop{\mathrm{argmax}}}}

\newcommand{\thetav}{\ensuremath{\bm{\theta}}}
\newcommand{\xv}{\ensuremath{\bm{x}}}
\newcommand{\yv}{\ensuremath{\bm{y}}}
\newcommand{\zv}{\ensuremath{\bm{z}}}

\newtheorem{lemma-ap}{Lemma}

\newtheorem{claim-ap}{Claim}

\makeatletter
\usepackage{xspace}
\def\@onedot{\ifx\@let@token.\else.\null\fi\xspace}
\DeclareRobustCommand\onedot{\futurelet\@let@token\@onedot}

\newcommand{\figref}[1]{Fig\onedot~\ref{#1}}

\newcommand{\secref}[1]{Sec\onedot~\ref{#1}}
\newcommand{\tabref}[1]{Table~\ref{#1}}

\newcommand{\by}[2]{\ensuremath{#1 \! \times \! #2}}

\def\eg{\emph{e.g}\onedot} 
\def\ie{\emph{i.e}\onedot} 
 
 \def\vs{\emph{vs}\onedot}
\def\wrt{w.r.t\onedot}

\makeatother

\begin{document} 

\title{Weakly- and Semi-Supervised Learning of a Deep Convolutional
Network for Semantic Image Segmentation}

\author{%
George Papandreou\thanks{The first two authors contributed equally to this work.}\\
Google, Inc.\\
{\tt\small gpapan@google.com}
\and
Liang-Chieh Chen$^*$\\
UCLA\\
{\tt\small lcchen@cs.ucla.edu}
\and
Kevin Murphy\\
Google, Inc.\\
{\tt\small kpmurphy@google.com}
\and
Alan L. Yuille \\
UCLA\\
{\tt\small yuille@stat.ucla.edu}
}

\maketitle

\begin{abstract}
Deep convolutional neural networks (DCNNs) trained on a large number
of images with strong pixel-level annotations have recently
significantly pushed the state-of-art in semantic image
segmentation. We study the more challenging problem of learning DCNNs
for semantic image segmentation from either (1) weakly annotated
training data such as bounding boxes or image-level labels or (2) a
combination of few strongly labeled and many weakly labeled images,
sourced from one or multiple datasets.
We develop Expectation-Maximization (EM) methods for semantic image
segmentation model training under these weakly supervised and
semi-supervised settings. Extensive experimental evaluation shows that
the proposed techniques can learn models delivering competitive
results on the challenging PASCAL VOC 2012 image segmentation
benchmark, while requiring significantly less annotation effort.
We share source code implementing the proposed system at
\url{https://bitbucket.org/deeplab/deeplab-public}. 
\end{abstract}


\section{Introduction}
\label{sec:intro}

Semantic image segmentation refers to the problem of assigning a
semantic label (such as ``person'', ``car'' or ``dog'') to every pixel
in the image. Various approaches have been tried over the years, but
according to the results  on the challenging Pascal VOC 2012
segmentation benchmark, the best performing methods all use some kind
of Deep Convolutional Neural Network (DCNN)
\cite{bell2014material, chen2014semantic, dai2014convolutional, 
  farabet2013learning, long2014fully, mostajabi2014feedforward,  
  zheng2015conditional}. 

In this paper, we work with the DeepLab-CRF approach of
\cite{chen2014semantic, zheng2015conditional}. This combines a DCNN
with a fully connected Conditional Random Field (CRF)
\cite{krahenbuhl2011efficient}, in order to get high resolution
segmentations. This model achieves state-of-art results on the
challenging PASCAL VOC segmentation benchmark
\cite{everingham2014pascal}, delivering a mean intersection-over-union
(IOU) score exceeding 70\%.

A key bottleneck in building this class of DCNN-based segmentation
models is that they typically require pixel-level annotated images
during training. Acquiring such data is an expensive,
time-consuming annotation effort. Weak annotations, in the form of
bounding boxes (\ie, coarse object locations) or image-level labels
(\ie, information about which object classes are present) are far
easier to collect than detailed pixel-level annotations. We
develop new methods for training DCNN image segmentation
models from weak annotations, either alone or in combination with a
small number of strong annotations. Extensive experiments, in
which we achieve performance up to 69.0\%, demonstrate the
effectiveness of the proposed techniques.

According to \cite{lin2014microsoft}, collecting bounding boxes around
each class instance in the image is about 15 times faster/cheaper than
labeling images at the pixel level. We demonstrate that it is possible
to learn a DeepLab-CRF model delivering 62.2\% IOU on the PASCAL VOC
2012 test set by training it on a simple foreground/background
segmentation of the bounding box annotations.

An even cheaper form of data to collect is image-level labels, which
specify the presence or absence of semantic classes, but not the
object locations. Most existing approaches for training semantic
segmentation models from this kind of very weak labels use multiple
instance learning (MIL) techniques. However, even recent
weakly-supervised methods such as \cite{long2014fully} deliver
significantly inferior results compared to their fully-supervised
counterparts, only achieving 25.7\%. Including additional trainable
objectness \cite{BingObj2014} or segmentation
\cite{arbelaez2014multiscale} modules that largely increase the
system complexity, \cite{Pinheiro2015} has improved performance to
40.6\%, which still significantly lags performance of fully-supervised
systems.

We develop novel online Expectation-Maximization (EM) methods for
training DCNN semantic segmentation models from weakly annotated
data. The proposed algorithms alternate between estimating the latent
pixel labels (subject to the weak annotation constraints), and
optimizing the DCNN parameters using stochastic gradient descent
(SGD). When we only have access to image-level annotated training data, we
achieve 39.6\%, close to \cite{Pinheiro2015} but without relying on
any external objectness or segmentation module. More importantly, our
EM approach also excels in the semi-supervised scenario which is very
important in practice. Having access to a small number of strongly
(pixel-level) annotated images and a large number of weakly (bounding
box or image-level) annotated images, the proposed algorithm can
almost match the performance of the fully-supervised system. For
example, having access to 2.9k pixel-level images and 9k image-level
annotated images yields 68.5\%, only 2\% inferior the performance of
the system trained with all 12k images strongly annotated at the pixel
level. Finally, we show that using additional weak or strong
annotations from the MS-COCO dataset can further improve results,
yielding 73.9\% on the PASCAL VOC 2012 benchmark.

\vspace{-0.5cm} 
\paragraph{Contributions}
In summary, our main contributions are:
\begin{enumerate}
\setlength{\itemsep}{0.5pt}
\item We present EM algorithms for training with image-level or
  bounding box annotation, applicable to both the weakly-supervised
  and semi-supervised settings.
\item We show that our approach achieves excellent performance when
  combining a small number of pixel-level annotated images with a
  large number of image-level or bounding box annotated images, nearly
  matching the results achieved when all training images have
  pixel-level annotations.
\item We show that combining weak or strong annotations across
  datasets yields further improvements. In particular, we reach 73.9\%
  IOU performance on PASCAL VOC 2012 by combining annotations from the
  PASCAL and MS-COCO datasets.
\end{enumerate}










\section{Related work}
\label{sec:related}

Training segmentation models with only image-level labels has been a
challenging problem in the literature \cite{duygulu2002object,
  verbeek2007region, vezhnevets2012weakly, xu2014tell}. Our work
is most related to other recent DCNN models such as
\cite{pathak2014fully, Pinheiro2015}, who also study the weakly
supervised setting. They both develop MIL-based algorithms for the
problem. In contrast, our model employs an EM algorithm, which
similarly to \cite{Lu2013sports} takes into account the weak labels
when inferring the latent image segmentations. Moreover,
\cite{Pinheiro2015} proposed to smooth the prediction results by
region proposal algorithms, \eg, CPMC \cite{carreira2012cpmc} and MCG \cite{arbelaez2014multiscale},
learned on pixel-segmented images. Neither \cite{pathak2014fully,
  Pinheiro2015} cover the semi-supervised setting.

Bounding box annotations have been utilized for semantic segmentation
by \cite{xia2013semantic, zhu2014learning}, while
\cite{guillaumin2014imagenet, kumar2011learning, xu2015learning} describe schemes exploiting both
image-level labels and bounding box annotations. \cite{chen2014beat} attained human-level
accuracy for car segmentation by using 3D bounding
boxes. Bounding box annotations are also commonly used in interactive
segmentation \cite{lempitsky2009image, rother2004grabcut}; 
we show that such foreground/background segmentation
methods can effectively estimate object segments accurate enough for
training a DCNN semantic segmentation system. Working in a setting
very similar to ours, \cite{dai2015boxsup} employed MCG
\cite{arbelaez2014multiscale} (which requires training from
pixel-level annotations) to infer object masks from bounding box
labels during DCNN training.



\section{Proposed Methods}
\label{sec:methods}

We build on the DeepLab model for semantic image segmentation proposed
in \cite{chen2014semantic}. This uses a DCNN to predict the label
distribution per pixel, followed by a fully-connected (dense) CRF
\cite{krahenbuhl2011efficient} to smooth the predictions while
preserving image edges. In this paper, we focus for simplicity on
methods for training the DCNN parameters from weak labels, only using
the CRF at test time. Additional gains can be obtained by integrated
end-to-end training of the DCNN and CRF parameters
\cite{zheng2015conditional,chen2014learning}.

\vspace{-0.4cm}
\paragraph{Notation} We denote by $\xv$ the image values and $\yv$ the
segmentation map. In particular, $y_m  \in \{0, \ldots, L\}$ is the
pixel label at position $m \in \{1,\ldots,M\}$, assuming that 
we have the background as well as $L$ possible foreground labels and
$M$ is the number of pixels. Note that these pixel-level labels may
not be visible in the training set. We encode the set of image-level
labels by $\zv$, with $z_l = 1$, if the $l$-th label is present
anywhere in the image, \ie, if $\sum_m [y_m = l] > 0$.

\subsection{Pixel-level annotations}
\label{sec:train_pixel}

\begin{figure}[tbp!]
  \centering
  \includegraphics[width=0.9\linewidth]{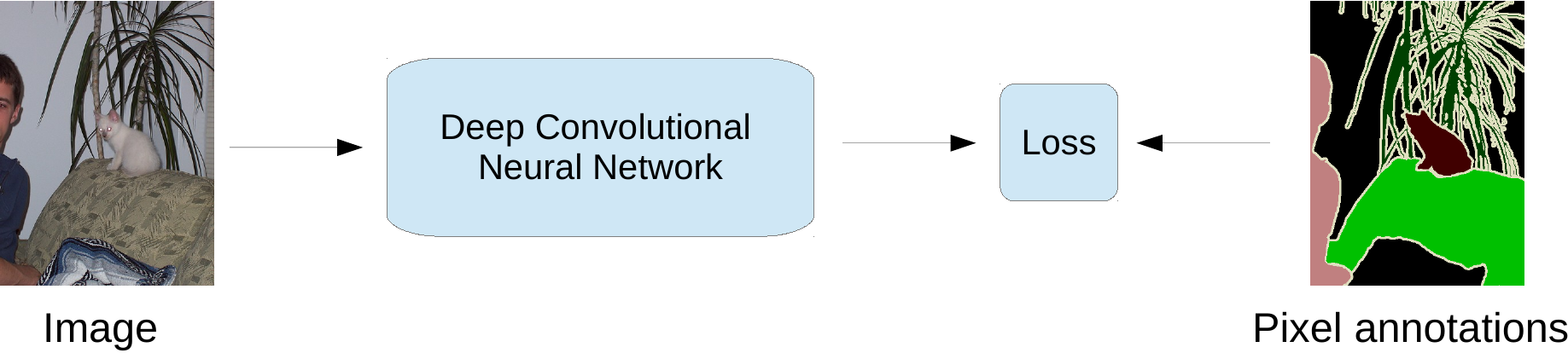} 
  \caption{DeepLab model training from fully annotated images.}
  \label{fig:model_train_pixel}
\end{figure}

In the fully supervised case illustrated in
\figref{fig:model_train_pixel}, the objective function is 
\begin{equation}
J(\thetav) = \log P(\yv | \xv; \thetav) = \sum_{m=1}^M \log P(y_m | \xv;
\thetav) \,,
\label{eqn:objStrong}
\end{equation}
where $\thetav$ is the vector of DCNN parameters. The per-pixel label
distributions are computed by
\begin{equation}
P(y_m | \xv ; \thetav) \propto \exp(f_m(y_m | \xv ; \thetav)) \,,
\label{eqn:pixel-prob}
\end{equation}
where $f_m(y_m | \xv; \thetav)$ is the output of the DCNN at pixel $m$.
We optimize $J(\thetav)$ by mini-batch SGD.

\subsection{Image-level annotations}
\label{sec:train_image}


When only image-level annotation is available, we can observe the
image values $\xv$ and the image-level labels $\zv$, but the
pixel-level segmentations $\yv$ are latent variables. We have the
following probabilistic graphical model:
\begin{equation}
P(\xv, \yv, \zv ; \thetav) = P(\xv) \left( \prod_{m=1}^M
P(y_m|\xv;\thetav) \right) P(\zv|\yv) \,.
\end{equation}

We pursue an EM-approach in order to learn the model parameters
$\thetav$ from training data. If we ignore terms that do not depend
on $\thetav$, the expected complete-data log-likelihood given the
previous parameter estimate $\thetav'$ is
\begin{equation}
Q(\thetav; \thetav')
= \sum_{\yv} P(\yv | \xv, \zv; \thetav') \log P(\yv | \xv; \thetav)
\approx \log P(\hat{\yv} | \xv; \thetav) \,,
\end{equation}
where we adopt a hard-EM approximation, estimating in the E-step of
the algorithm the latent segmentation by
\begin{eqnarray}
  \hat{\yv} &=& \argmax_{\yv} P(\yv | \xv; \thetav') P(\zv | \yv) \\
  &=& \argmax_{\yv} \log P(\yv | \xv; \thetav') + \log P(\zv | \yv) \\
  &=& \argmax_{\yv} \left( \sum_{m=1}^M f_m(y_m | \xv ; \thetav') +
  \log P(\zv | \yv) \right)
 \,.
  \label{eqn:E-step}
\end{eqnarray}
In the M-step of the algorithm, we optimize $Q(\thetav; \thetav')
\approx \log P(\hat{\yv} | \xv; \thetav)$ by mini-batch SGD similarly
to \eqref{eqn:objStrong}, treating $\hat{\yv}$ as ground truth
segmentation.

To completely identify the E-step \eqref{eqn:E-step}, we need to
specify the observation model $P(\zv | \yv)$. We have experimented
with two variants, \textsl{EM-Fixed} and \textsl{EM-Adapt}.

\begin{algorithm}[tbp!]
  \centering
  \begin{algorithmic}[1]
    \algrenewcommand\algorithmicrequire{\textbf{Input:}}
    \Require Initial CNN parameters $\thetav'$,  potential parameters
    $b_{l}$, $l \in \{0,\dots,L\}$, image $\xv$, image-level label set
    $\zv$. 
    \algrenewcommand\algorithmicrequire{\textbf{E-Step:}}
    \Require For each image position $m$
    \State $\hat{f}_m (l) = f_m(l | \xv; \thetav') + b_l$, if $z_{l}=1$
    \State $\hat{f}_m (l) = f_m(l | \xv; \thetav')$, if $z_{l}=0$
    \State $\hat{y}_m = \argmax_{l} \hat{f}_m (l)$
    \algrenewcommand\algorithmicrequire{\textbf{M-Step:}}
    \Require
    \State $Q(\thetav;\thetav') = \log P(\hat{\yv} | \xv, \thetav) = \sum_{m = 1}^M \log P(\hat{y}_m | \xv, \thetav)$ 
    \State Compute $\nabla_{\thetav}Q(\thetav;\thetav')$ and use SGD to update $\thetav'$.
    \end{algorithmic}
  \caption{Weakly-Supervised EM (fixed bias version)}
  \label{algo:em_fixed_bias}
\end{algorithm}

\begin{figure}[tbp!]
  \centering
  \includegraphics[width=0.9\linewidth]{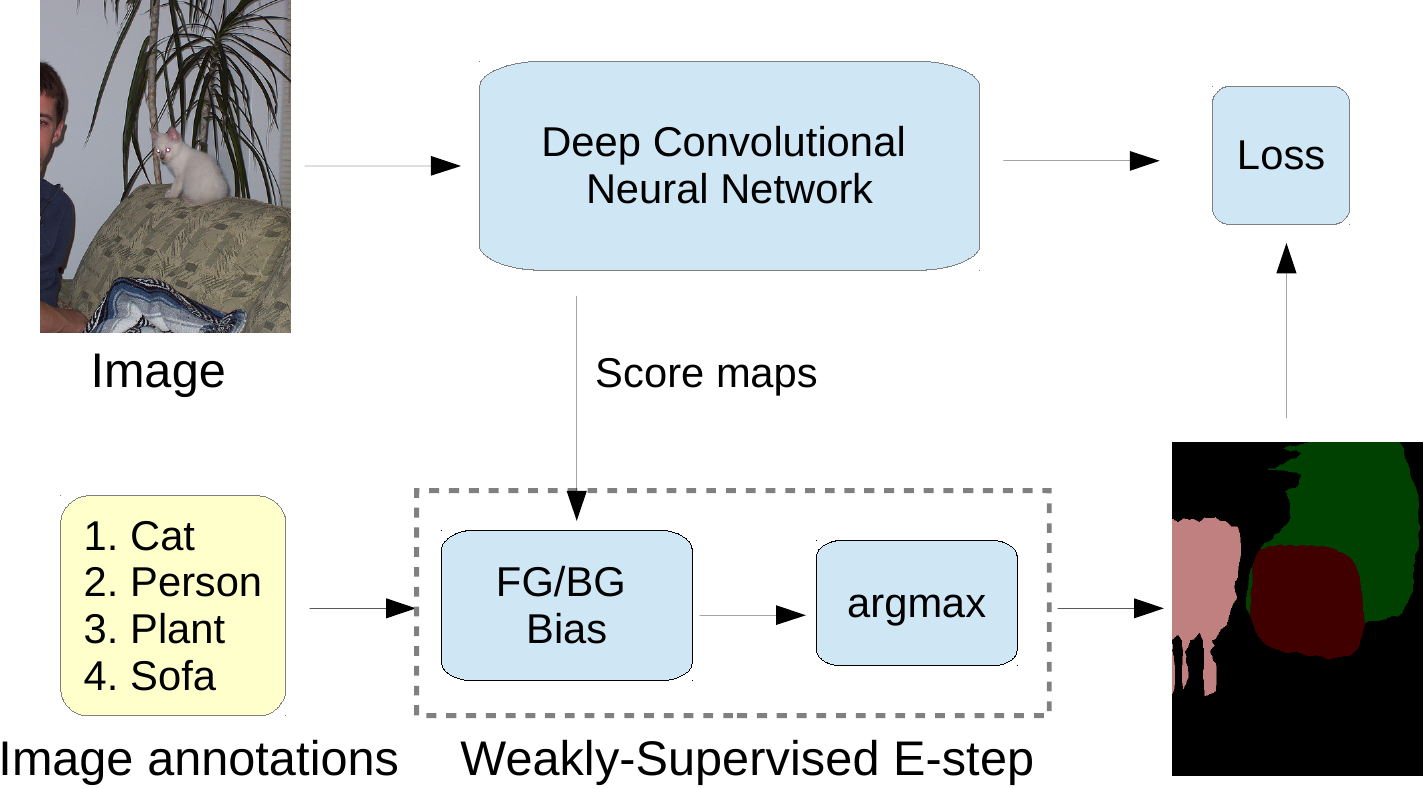} 
  \caption{DeepLab model training using image-level labels.}
  \label{fig:model_train_image}
\end{figure}

\vspace{-0.4cm}
\paragraph{EM-Fixed} In this variant, we assume that $\log P(\zv | \yv)$
factorizes over pixel positions as
\begin{equation}
  \log P(\zv | \yv) = \sum_{m=1}^M \phi(y_m,\zv) + \mathrm{(const)}  \,,
  \label{eqn:E-fixed}
\end{equation}
allowing us to estimate the E-step segmentation at each pixel separately
\begin{equation}
  \hat{y}_m = \argmax_{y_m} \hat{f}_m(y_m) \doteq f_m(y_m | \xv ; \thetav') + \phi(y_m,\zv) \,.
  \label{eqn:E-fixed_factor}
\end{equation}
We assume that
\begin{eqnarray}
  \phi(y_m = l,\zv) 
  &=& \left\{
  \begin{array}{ll}
    b_l &\mbox{if $z_{l}=1$} \\
    0   &\mbox{if $z_{l}=0$} \\
  \end{array}
  \right.
\end{eqnarray}
We set the parameters $b_l = b_{\mathrm{fg}}$, if $l > 0$ and $b_0 = b_{\mathrm{bg}}$,
with $b_{\mathrm{fg}} > b_{\mathrm{bg}} > 0$. Intuitively, this potential encourages a
pixel to be assigned to one of the image-level labels $\zv$. We choose
$b_{\mathrm{fg}} > b_{\mathrm{bg}}$, boosting present foreground classes more than the
background, to encourage full object coverage and avoid a degenerate
solution of all pixels being assigned to background. The procedure is
summarized in Algorithm~\ref{algo:em_fixed_bias} and illustrated in
\figref{fig:model_train_image}. 

\vspace{-0.4cm}
\paragraph{EM-Adapt}
In this method, we assume that $\log P(\zv | \yv) = \phi(\yv, \zv) +
\mathrm{(const)}$, where $\phi(\yv, \zv)$ takes the form of a
cardinality potential \cite{Li2014high, pletscher2012learning,
tarlow2012fast}. In particular, we encourage \emph{at least} a $\rho_l$
portion of the image area to be assigned to class $l$, if $z_l = 1$,
and enforce that no pixel is assigned to class $l$, if $z_l = 0$. We
set the parameters $\rho_l = \rho_{\mathrm{fg}}$, if $l > 0$ and
$\rho_0 = \rho_{\mathrm{bg}}$. Similar constraints
appear in \cite{delong2012fast, kuck2005individuals}. 

In practice, we employ a variant of Algorithm~\ref{algo:em_fixed_bias}.
We adaptively set the image- and class-dependent biases $b_l$
so as the prescribed proportion of the image area is assigned to the
background or foreground object classes. This acts as a powerful
constraint that explicitly prevents the background score from
prevailing in the whole image, also promoting higher foreground object
coverage. The detailed algorithm is described in the supplementary
material.

\vspace{-0.4cm}
\paragraph{EM vs. MIL}

It is instructive to compare our EM-based approach with two recent
Multiple Instance Learning (MIL) methods for learning semantic image
segmentation models \cite{pathak2014fully,Pinheiro2015}. The method in
\cite{pathak2014fully} defines an MIL classification objective based
on the per-class spatial maximum of the local label distributions of
\eqref{eqn:pixel-prob}, $\hat{P}(l | \xv ; \thetav) \doteq \max_m
P(y_m = l| \xv ; \thetav)$, and \cite{Pinheiro2015} adopts a softmax 
function. While this approach has worked well for
image classification tasks
\cite{oquab2014weakly,papandreou2014untangling}, it is less suited for
segmentation as it does not promote full object coverage: The
DCNN becomes tuned to focus on the most distinctive object
parts (\eg, human face) instead of capturing the whole object (\eg, human body).

\subsection{Bounding Box Annotations}
\label{sec:train_bbox}

\begin{figure}[tbp!]
  \centering
  \includegraphics[width=0.9\linewidth]{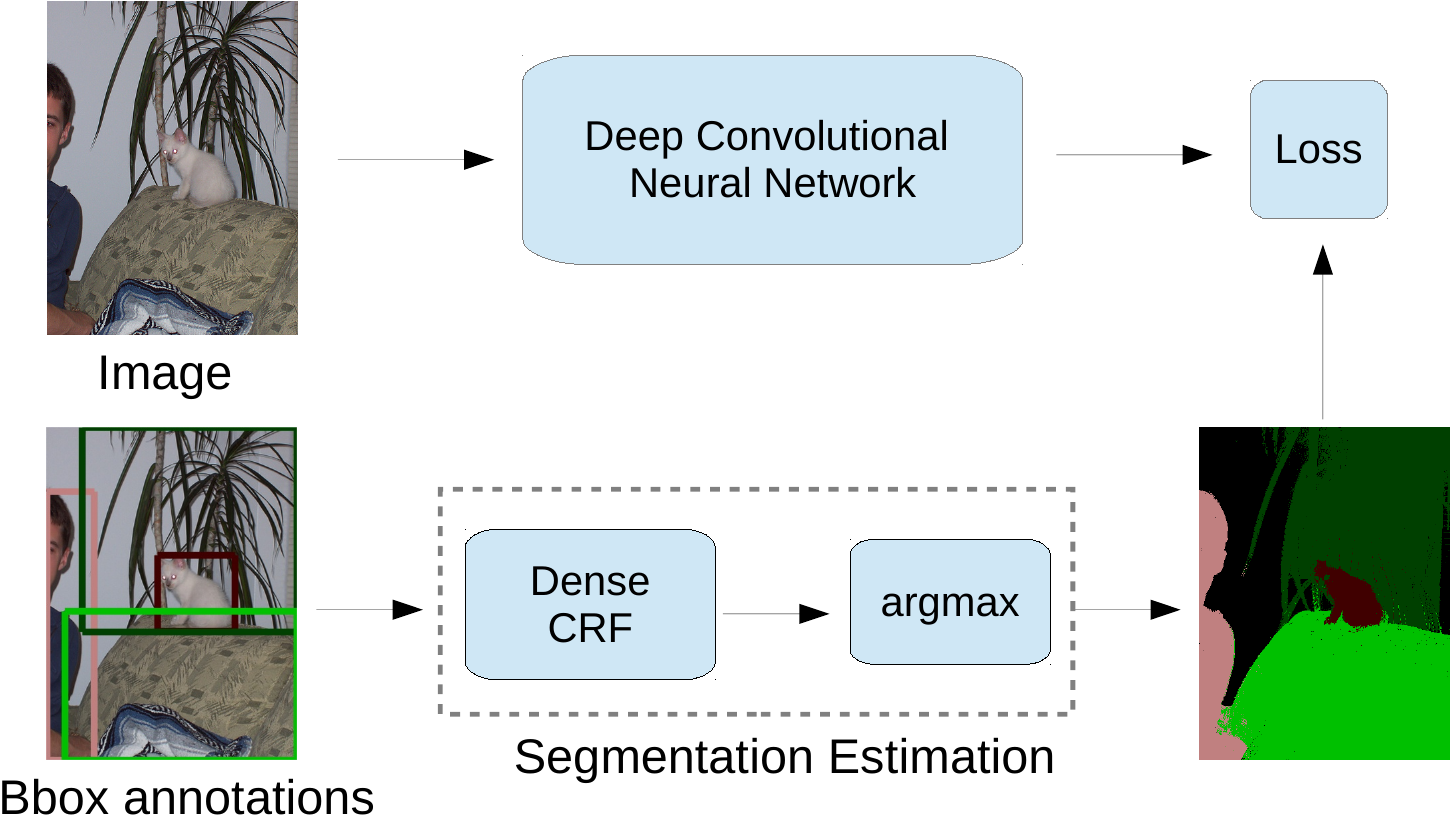} 
  \caption{DeepLab model training from bounding boxes.}
  \label{fig:model_train_bbox}
\end{figure}

We explore three alternative methods for training our
segmentation model from labeled bounding boxes.

The first \textsl{Bbox-Rect} method amounts to simply considering
each pixel within the bounding box as positive example for the
respective object class. Ambiguities are resolved by assigning pixels
that belong to multiple bounding boxes to the one that has the
smallest area.

\begin{figure}
  \centering
  \begin{tabular}{c c c c}
    \includegraphics[width=0.21\linewidth]{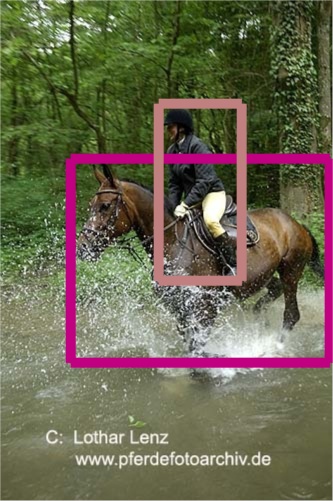} & 
    \includegraphics[width=0.21\linewidth]{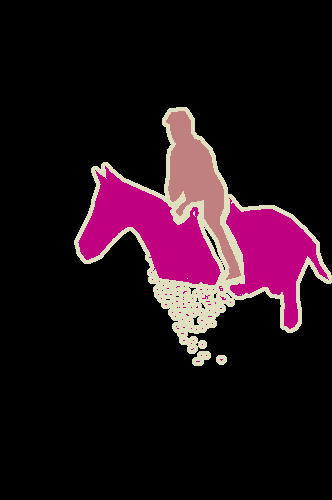} & 
    \includegraphics[width=0.21\linewidth]{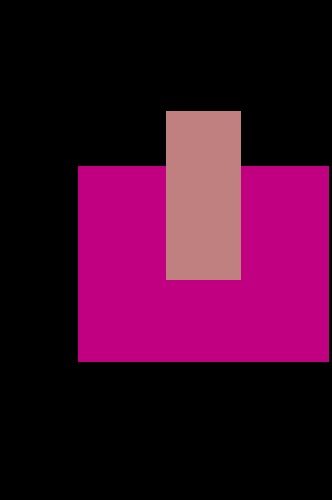} & 
    \includegraphics[width=0.21\linewidth]{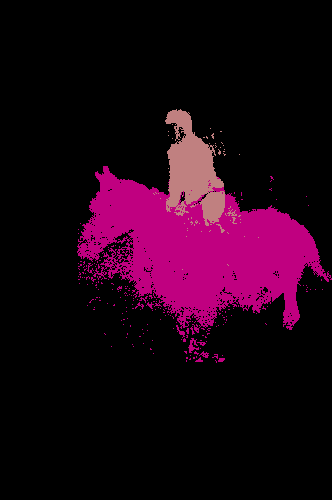} \\    
    {\scriptsize Image with Bbox} & {\scriptsize Ground-Truth} & {\scriptsize Bbox-Rect} & {\scriptsize Bbox-Seg}
  \end{tabular}
  \caption{Estimated segmentation from bounding box annotation.}
  \label{fig:bbox_illustration}
\end{figure}

The bounding boxes fully surround objects but also contain background
pixels that contaminate the training set with false positive examples
for the respective object classes. To filter out these background
pixels, we have also explored a second \textsl{Bbox-Seg} method in
which we perform automatic foreground/background segmentation.
To perform this segmentation, we use the same CRF as in DeepLab.
More specifically, we constrain the center area of the bounding box
($\alpha\%$ of pixels within the box) to be foreground, while we
constrain pixels outside the bounding box to be background. We
implement this by appropriately setting the unary terms of the CRF.
We then infer the labels for pixels in between. We cross-validate the
CRF parameters to maximize segmentation accuracy in a small
held-out set of fully-annotated images. This approach is similar to
the grabcut method of \cite{rother2004grabcut}. Examples of estimated
segmentations with the two methods are shown in
\figref{fig:bbox_illustration}.

The two methods above, illustrated in \figref{fig:model_train_bbox},
estimate segmentation maps from the bounding box annotation as a
pre-processing step, then employ the training procedure of
\secref{sec:train_pixel}, treating these estimated labels as
ground-truth.

Our third \textsl{Bbox-EM-Fixed} method is an EM algorithm that allows
us to refine the estimated segmentation maps throughout training. The
method is a variant of the \textsl{EM-Fixed} algorithm in
\secref{sec:train_image}, in which we boost the present foreground
object scores only within the bounding box area.

\subsection{Mixed strong and weak annotations}
\label{sec:train_semi}

\begin{figure}[tbp!]
  \centering
  \includegraphics[width=0.8\linewidth]{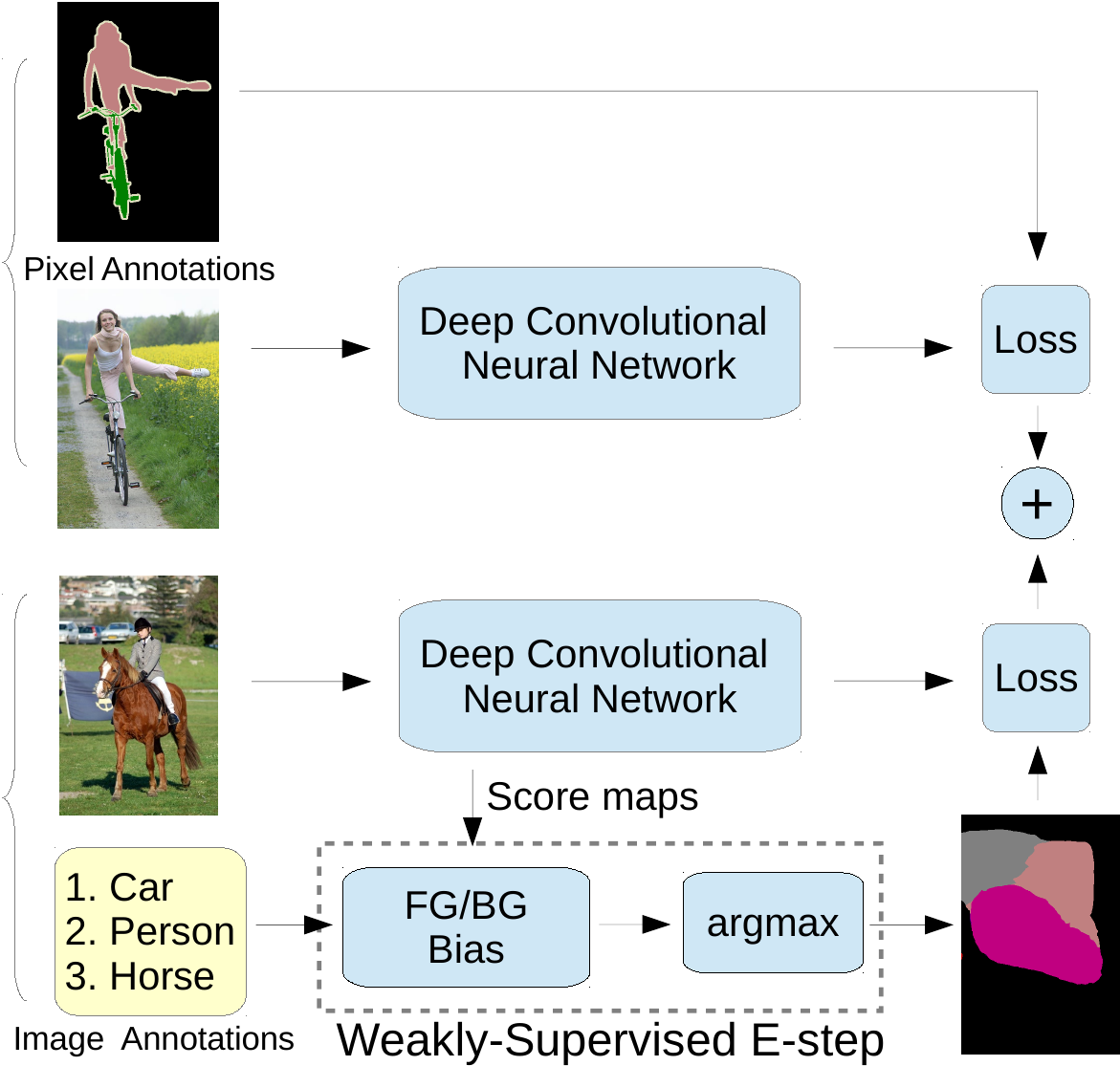} 
  \caption{DeepLab model training on a union of full (strong labels) 
    and image-level (weak labels) annotations.}
  \label{fig:model_illustrations_twoEnd}
\end{figure}

In practice, we often have access to a large number of weakly
image-level annotated images and can only afford to procure detailed
pixel-level annotations for a small fraction of these images. We
handle this hybrid training scenario by combining the methods
presented in the previous sections, as illustrated in
Figure~\ref{fig:model_illustrations_twoEnd}. In SGD training of our
deep CNN models, we bundle to each mini-batch a fixed proportion of
strongly/weakly annotated images, and employ our EM algorithm in estimating
at each iteration the latent semantic segmentations for the weakly
annotated images. 

\section{Experimental Evaluation}
\label{sec:experiments}

\subsection{Experimental Protocol}
\vspace{-0cm}
\paragraph{Datasets} 
\label{sec:dataset}

The proposed training methods are evaluated on the PASCAL
VOC 2012 segmentation benchmark \cite{everingham2014pascal},
consisting of 20 foreground object classes and one background
class. The segmentation part of the original PASCAL VOC 2012 dataset contains
$1464$ (\textsl{train}), $1449$ (\textsl{val}), and $1456$
(\textsl{test}) images for training, validation, and test,
respectively. We also use the extra annotations
provided by \cite{hariharan2011semantic}, resulting in augmented sets
of $10,582$ (\textsl{train\_aug}) and $12,031$
(\textsl{trainval\_aug}) images. We have also experimented with the
large MS-COCO 2014 dataset \cite{lin2014microsoft}, which contains
$123,287$ images in its \textsl{trainval} set. The MS-COCO 2014
dataset has $80$ foreground object classes and one background class
and is also annotated at the pixel level.

The performance is measured in terms of pixel intersection-over-union
(IOU) averaged across the 21 classes. We first evaluate our proposed
methods on the PASCAL VOC 2012 \textsl{val} set. We then report
our results on the official PASCAL VOC 2012 benchmark \textsl{test}
set (whose annotations are not released). We also compare our
\textsl{test} set results with other competing methods.

\vspace{-0.4cm}
\paragraph{Reproducibility} We have implemented the proposed methods
by extending the excellent Caffe framework \cite{jia2014caffe}. We
share our source code, configuration files, and trained models that
allow reproducing the results in this paper at a companion web site
\url{https://bitbucket.org/deeplab/deeplab-public}.

\vspace{-0.4cm}
\paragraph{Weak annotations}
In order to simulate the situations where only weak annotations are
available and to have fair comparisons (e.g., use the same images for
all settings), we generate the weak annotations from the pixel-level
annotations. The image-level labels are easily generated by
summarizing the pixel-level annotations, while the bounding box
annotations are produced by drawing rectangles tightly containing each
object instance (PASCAL VOC 2012 also provides instance-level
annotations) in the dataset.

\vspace{-0.4cm}
\paragraph{Network architectures}
We have experimented with the two DCNN architectures of
\cite{chen2014semantic}, with parameters initialized from the VGG-16
ImageNet \cite{deng2009imagenet} pretrained model of \cite{simonyan2014very}. They differ in
the receptive field of view (FOV) size. We have found that large FOV
(\by{224}{224}) performs best when at least some training images are
annotated at the pixel level, whereas small FOV (\by{128}{128})
performs better when only image-level annotations are available. In
the main paper we report the results of the best architecture for each
setup and defer the full comparison between the two FOVs to the
supplementary material.

\vspace{-0.4cm}
\paragraph{Training}
\label{sec:resultsTraining}
We employ our proposed training methods to learn the DCNN component of
the DeepLab-CRF model of \cite{chen2014semantic}. For SGD, we
use a mini-batch of 20-30 images and initial learning rate of $0.001$
($0.01$ for the final classifier layer), multiplying the learning rate
by 0.1 after a fixed number of iterations. We use momentum of $0.9$
and a weight decay of $0.0005$. Fine-tuning our network on PASCAL VOC
2012 takes about 12 hours on a NVIDIA Tesla K40 GPU. 

Similarly to \cite{chen2014semantic}, we decouple the DCNN and Dense
CRF training stages and learn the CRF parameters by cross validation
to maximize IOU segmentation accuracy in a held-out set of 100
Pascal \textsl{val} fully-annotated images. We use 10 mean-field
iterations for Dense CRF inference \cite{krahenbuhl2011efficient}.
Note that the IOU scores are typically 3-5\% worse if we don't use
the CRF for post-processing of the results.

\subsection{Pixel-level annotations}
\label{sec:test_pixel}

We have first reproduced the results of
\cite{chen2014semantic}. Training the DeepLab-CRF model with strong
pixel-level annotations on PASCAL VOC 2012, we achieve a mean IOU
score of 67.6\% on \textsl{val} and 70.3\% on \textsl{test}; see
method \textsl{DeepLab-CRF-LargeFOV} in
\cite[Table~1]{chen2014semantic}.

\subsection{Image-level annotations}
\label{sec:test_image}

\begin{table}[tbp!]
  \centering
  \scalebox{0.70}{
  \begin{tabular}{|c|c|c|c|c|}
    \hline
    Method                   & \#Strong & \#Weak  & val IOU\\
    \hline
    EM-Fixed (Weak)          & -        &  10,582 & 20.8\\
    EM-Adapt (Weak)          & -        &  10,582 & 38.2\\
    \hline
    \multirow{6}{*}{EM-Fixed (Semi)}
                             & 200      &  10,382 & 47.6\\
                             & 500      &  10,082 & 56.9\\
                             & 750      &  9,832  & 59.8\\
                             & 1,000    &  9,582  & 62.0\\
                             & 1,464    &  5,000  &  63.2\\
                             & 1,464    &  9,118  &  64.6\\
    \hline
    \multirow{2}{*}{Strong}  & 1,464    &  -      &  62.5\\
                             & 10,582   &  -      &  67.6\\
    \hline
  \end{tabular}  
  }
  \caption{VOC 2012 \textsl{val} performance for varying number of
    pixel-level (strong) and image-level (weak) annotations
    (\secref{sec:test_image}).}
  \label{tab:image_annot}
\end{table}



\begin{table}[t]
  \centering
  \scalebox{0.70}{
  \begin{tabular}{|c|c|c|c|}
    \hline
    Method                          & \#Strong & \#Weak & test IOU \\
    \hline
    MIL-FCN \cite{pathak2014fully}  &     -    &   10k  &  25.7 \\
    MIL-sppxl \cite{Pinheiro2015}   &     -    &   760k &  35.8 \\
    MIL-obj \cite{Pinheiro2015}     &   BING   &   760k &  37.0 \\
    MIL-seg \cite{Pinheiro2015}     &    MCG   &   760k &  40.6 \\
    \hline\hline
    EM-Adapt (Weak)                 &     -    &   12k  &  39.6 \\
    \hline
    \multirow{2}{*}{EM-Fixed (Semi)}&    1.4k  &   10k  &  66.2 \\
                                    &    2.9k  &    9k  &  68.5 \\
    \hline \hline
    Strong \cite{chen2014semantic}  &    12k   &    -   &  70.3 \\
    \hline
  \end{tabular}
  }
  \caption{VOC 2012 \textsl{test} performance for varying number of
    pixel-level (strong) and image-level (weak) annotations
    (\secref{sec:test_image}).}
  \label{tab:image_annot_test}
\end{table}

\vspace{-0cm}
\paragraph{Validation results}
We evaluate our proposed methods in training the
DeepLab-CRF model using image-level weak annotations from
the 10,582 PASCAL VOC 2012 \textsl{train\_aug} set, generated as
described in \secref{sec:dataset} above. We report the \textsl{val}
performance of our two weakly-supervised EM variants described in
\secref{sec:train_image}. In the \textsl{EM-Fixed} variant we use
$b_{\mathrm{fg}} = 5$ and $b_{\mathrm{bg}} = 3$ as fixed foreground and background
biases. We found the results to be quite sensitive to the difference
$b_{\mathrm{fg}} - b_{\mathrm{bg}}$ but not very sensitive to their absolute values.
In the adaptive \textsl{EM-Adapt} variant we constrain at least
$\rho_{\mathrm{bg}} = 40\%$ of the image area to be assigned to background and
at least $\rho_{\mathrm{fg}} = 20\%$ of the image area to be assigned to
foreground (as specified by the weak label set).

We also examine using weak image-level annotations in addition to
a varying number of pixel-level annotations, within the semi-supervised
learning scheme of \secref{sec:train_semi}. In this \textsl{Semi} setting we employ
strong annotations of a subset of PASCAL VOC 2012 \textsl{train} set
and use the weak image-level labels from another non-overlapping
subset of the \textsl{train\_aug} set. We perform segmentation
inference for the images that only have image-level labels by means of
\textsl{EM-Fixed}, which we have found to perform better than
\textsl{EM-Adapt} in the semi-supervised training setting.

The results are summarized in Table~\ref{tab:image_annot}. We see that
the EM-Adapt algorithm works much better than the EM-Fixed algorithm
when we only have access to image level annotations, 20.8\% vs. 38.2\%
validation IOU. Using 1,464 pixel-level and 9,118 image-level
annotations in the EM-Fixed semi-supervised setting significantly
improves performance, yielding 64.6\%. Note that image-level
annotations are helpful, as training only with the 1,464 pixel-level
annotations only yields 62.5\%.

\vspace{-0.4cm}
\paragraph{Test results}
In Table~\ref{tab:image_annot_test} we report our \textsl{test}
results. We compare the proposed methods with the recent MIL-based
approaches of \cite{pathak2014fully, Pinheiro2015}, which also report
results obtained with image-level annotations on the VOC
benchmark. Our EM-Adapt method yields 39.6\%, which improves over
MIL-FCN \cite{pathak2014fully} by a large 13.9\% margin. As
\cite{Pinheiro2015} shows, MIL can become more competitive if
additional segmentation information is introduced: Using low-level
superpixels, MIL-sppxl \cite{Pinheiro2015} yields 35.8\% and is still
inferior to our EM algorithm. Only if augmented with BING
\cite{BingObj2014} or MCG \cite{arbelaez2014multiscale} can
MIL obtain results comparable to ours (MIL-obj: 37.0\%, MIL-seg:
40.6\%) \cite{Pinheiro2015}. Note, however, that both BING and MCG
have been trained with bounding box or pixel-annotated data on the
PASCAL \textsl{train} set, and thus both MIL-obj and MIL-seg
indirectly rely on bounding box or pixel-level PASCAL annotations.

The more interesting finding of this experiment is that including very
few strongly annotated images in the semi-supervised setting
significantly improves the performance compared to the pure
weakly-supervised baseline. For example, using 2.9k pixel-level
annotations along with 9k image-level annotations in the
semi-supervised setting yields 68.5\%. We would like to highlight that
this result surpasses all techniques which are not based on the
DCNN+CRF pipeline of \cite{chen2014semantic} (see
\tabref{tab:cross_data_test}), even if trained with all available
pixel-level annotations.

\subsection{Bounding box annotations}
\label{sec:test_bbox}
\vspace{-0cm}
\paragraph{Validation results}

In this experiment, we train the DeepLab-CRF model using bounding box
annotations from the \textsl{train\_aug} set. We estimate the
training set segmentations in a pre-processing step using the
\textsl{Bbox-Rect} and \textsl{Bbox-Seg} methods described in
\secref{sec:train_bbox}. We assume that we also have access to 100
fully-annotated PASCAL VOC 2012 \textsl{val} images which we have used
to cross-validate the value of the single \textsl{Bbox-Seg} parameter
$\alpha$ (percentage of the center bounding box area constrained to be
foreground). We varied $\alpha$ from 20\% to 80\%, finding that
$\alpha = 20\%$ maximizes accuracy in terms of IOU in recovering
the ground truth foreground from the bounding box. We also examine the
effect of combining these weak bounding box annotations with strong
pixel-level annotations, using the semi-supervised learning methods of
\secref{sec:train_semi}.

The results are summarized in Table~\ref{tab:bbox_annot}. When using
only bounding box annotations, we see that \textsl{Bbox-Seg} improves
over \textsl{Bbox-Rect} by 8.1\%, and gets within 7.0\% of the
strong pixel-level annotation result. We observe that combining 1,464
strong pixel-level annotations with weak bounding box annotations
yields 65.1\%, only 2.5\% worse than the strong pixel-level annotation
result. In the semi-supervised learning settings and 1,464 strong
annotations, \textsl{Semi-Bbox-EM-Fixed} and \textsl{Semi-Bbox-Seg}
perform similarly.

\begin{table}[tbp!]
  \centering
  \scalebox{0.70}{
  \begin{tabular}{|c|c|c|c|}
    \hline
    Method                 & \#Strong & \#Box   & val IOU\\
    \hline
    Bbox-Rect (Weak)       & -        &  10,582 & 52.5\\
    Bbox-EM-Fixed (Weak)   & -        &  10,582 & 54.1\\
    Bbox-Seg (Weak)        & -        &  10,582 & 60.6\\
    \hline
    Bbox-Rect (Semi)       & 1,464    &  9,118  & 62.1\\
    Bbox-EM-Fixed (Semi)   & 1,464    &  9,118  & 64.8\\
    Bbox-Seg (Semi)        & 1,464    &  9,118  & 65.1\\
    \hline
    \multirow{2}{*}{Strong}& 1,464    &  -      & 62.5\\
                           & 10,582   &  -      & 67.6\\
    \hline
  \end{tabular}  
  }
  \caption{VOC 2012 \textsl{val} performance for varying number of
    pixel-level (strong) and bounding box (weak) annotations
    (\secref{sec:test_bbox}).}
  \label{tab:bbox_annot}
\end{table}

\begin{table}[t]
  \centering
  \scalebox{0.70}{
  \begin{tabular}{|c|c|c|c|}
    \hline
    Method                      & \#Strong    & \#Box & test IOU\\
    \hline
    BoxSup \cite{dai2015boxsup} & MCG         & 10k   & 64.6\\
    BoxSup \cite{dai2015boxsup} & 1.4k (+MCG) & 9k    & 66.2\\
   \hline \hline
    Bbox-Rect (Weak)            & -           & 12k   & 54.2\\
    Bbox-Seg (Weak)             & -           & 12k   & 62.2\\
    \hline
    Bbox-Seg (Semi)             & 1.4k        & 10k   & 66.6\\ 
    Bbox-EM-Fixed (Semi)        & 1.4k        & 10k   & 66.6\\
    Bbox-Seg (Semi)             & 2.9k        &  9k   & 68.0\\ 
    Bbox-EM-Fixed (Semi)        & 2.9k        &  9k   & 69.0\\
    \hline \hline
    Strong \cite{chen2014semantic} &    12k   &    -  & 70.3\\
    \hline
  \end{tabular}
  }
  \caption{VOC 2012 \textsl{test} performance for varying number of
    pixel-level (strong) and bounding box (weak) annotations
    (\secref{sec:test_bbox}).}
  \label{tab:bbox_annot_test}
\end{table}

\vspace{-0.4cm}
\paragraph{Test results}

In Table~\ref{tab:bbox_annot_test} we report our \textsl{test}
results. We compare the proposed methods with the very recent BoxSup
approach of \cite{dai2015boxsup}, which also uses
bounding box annotations on the VOC 2012 segmentation
benchmark. Comparing our alternative Bbox-Rect (54.2\%) and Bbox-Seg
(62.2\%) methods, we see that simple foreground-background
segmentation provides much better segmentation masks for DCNN training
than using the raw bounding boxes. BoxSup does 2.4\% better, however
it employs the MCG segmentation proposal mechanism
\cite{arbelaez2014multiscale}, which has been trained with
pixel-annotated data on the PASCAL \textsl{train} set; it thus
indirectly relies on pixel-level annotations.

When we also have access to pixel-level annotated images, our
performance improves to 66.6\% (1.4k strong annotations) or 69.0\%
(2.9k strong annotations). In this semi-supervised setting we
outperform BoxSup (66.6\% \vs 66.2\% with 1.4k strong annotations),
although we do not use MCG. Interestingly, Bbox-EM-Fixed improves over
Bbox-Seg as we add more strong annotations, and it performs
1.0\% better (69.0\% \vs 68.0\%) with 2.9k strong annotations. This
shows that the E-step of our EM algorithm can estimate the object
masks better than the foreground-background segmentation
pre-processing step when enough pixel-level annotated images are
available.

Comparing with \secref{sec:test_image}, note that 2.9k strong
+ 9k image-level annotations yield 68.5\%
(\tabref{tab:image_annot_test}), while 2.9k strong + 9k bounding box
annotations yield 69.0\% (\tabref{tab:bbox_annot}). This finding
suggests that bounding box annotations add little value over
image-level annotations when a sufficient number of pixel-level
annotations is also available.

\subsection{Exploiting Annotations Across Datasets}
\label{sec:cross_dataset}
\vspace{-0cm}
\paragraph{Validation results}

We present experiments leveraging the 81-label MS-COCO
dataset as an additional source of data in learning the DeepLab model
for the 21-label PASCAL VOC 2012 segmentation task. We consider
three scenarios:
\begin{itemize}
\vspace{-0.2cm}
\setlength{\itemsep}{0.5pt}
\item \textsl{Cross-Pretrain (Strong)}: Pre-train DeepLab on MS-COCO,
  then replace the top-level network weights and fine-tune on
  Pascal VOC 2012, using pixel-level annotation in both datasets.
\item \textsl{Cross-Joint (Strong)}: Jointly train DeepLab on Pascal VOC
  2012 and MS-COCO, sharing the top-level network weights for the
  common classes, using pixel-level annotation in both datasets.
\item \textsl{Cross-Joint (Semi)}: Jointly train DeepLab on Pascal VOC
  2012 and MS-COCO, sharing the top-level network weights for the
  common classes, using the pixel-level labels from PASCAL and
  varying the number of pixel- and image-level labels from MS-COCO.
\end{itemize}
\vspace{-0.2cm}
In all cases we use strong pixel-level annotations for all 10,582
\textsl{train\_aug} PASCAL images.

\begin{table}[t]
  \centering
  \scalebox{0.70}{
  \begin{tabular}{|c|c|c|c|c|}
    \hline
    Method                  & \#Strong COCO & \#Weak COCO & val IOU \\
    \hline
    PASCAL-only             & -        & -       &  67.6 \\
    \hline
    EM-Fixed (Semi)         & -        & 123,287 &  67.7 \\
    \hline
    Cross-Joint (Semi)      & 5,000    & 118,287 &  70.0 \\
    \hline
    Cross-Joint (Strong)    & 5,000    &  -      &  68.7 \\
    Cross-Pretrain (Strong) & 123,287  &  -      &  71.0 \\
    Cross-Joint (Strong)    & 123,287  &  -      &  71.7 \\
    \hline
  \end{tabular}  
  }
  \caption{VOC 2012 \textsl{val} performance using strong annotations
    for all 10,582 \textsl{train\_aug} PASCAL images and a varying
    number of strong and weak MS-COCO annotations
    (\secref{sec:cross_dataset}).}
  \label{tab:cross_data}
\end{table}

\begin{table}[t]
  \centering
  \scalebox{0.70}{
  \begin{tabular}{|c|c|}
    \hline
    Method & test IOU\\
    \hline \hline
    MSRA-CFM \cite{dai2014convolutional}               & 61.8 \\
    FCN-8s \cite{long2014fully}                        & 62.2 \\
    Hypercolumn \cite{hariharan2014hypercolumns}       & 62.6 \\
    TTI-Zoomout-16 \cite{mostajabi2014feedforward}     & 64.4 \\
    \hline
    DeepLab-CRF-LargeFOV \cite{chen2014semantic}       & 70.3 \\
    BoxSup (Semi, with weak COCO) \cite{dai2015boxsup} & 71.0 \\
    DeepLab-CRF-LargeFOV (Multi-scale net) \cite{chen2014semantic} & 71.6 \\
    Oxford\_TVG\_CRF\_RNN\_VOC \cite{zheng2015conditional} & 72.0 \\
    Oxford\_TVG\_CRF\_RNN\_COCO \cite{zheng2015conditional} & 74.7 \\
    \hline \hline
    Cross-Pretrain (Strong)                            & 72.7 \\
    Cross-Joint (Strong)                               & 73.0 \\
    Cross-Pretrain (Strong, Multi-scale net)           & 73.6 \\
    Cross-Joint (Strong, Multi-scale net)              & 73.9 \\
    \hline
  \end{tabular}
  }
  \caption{VOC 2012 \textsl{test} performance using PASCAL
    and MS-COCO annotations (\secref{sec:cross_dataset}).}
  \label{tab:cross_data_test}
\end{table}

We report our results on the PASCAL VOC 2012 \textsl{val} in
\tabref{tab:cross_data}, also including for comparison our best
PASCAL-only 67.6\% result exploiting all 10,582 strong annotations as
a baseline. When we employ the weak MS-COCO annotations
(\textsl{EM-Fixed (Semi)}) we obtain 67.7\% IOU, which does not
improve over the PASCAL-only baseline. However, using strong labels from
5,000 MS-COCO images (4.0\% of the MS-COCO dataset) and weak labels
from the remaining MS-COCO images in the \textsl{Cross-Joint (Semi)}
semi-supervised scenario yields 70.0\%, a significant 2.4\% boost over
the baseline. This \textsl{Cross-Joint (Semi)} result is also 1.3\%
better than the 68.7\% performance obtained using only the 5,000
strong and no weak annotations from MS-COCO. As expected, our best
results are obtained by using all 123,287 strong MS-COCO
annotations, 71.0\% for \textsl{Cross-Pretrain (Strong)} and
71.7\% for \textsl{Cross-Joint (Strong)}. We observe
that cross-dataset augmentation improves by 4.1\% over the best
PASCAL-only result. Using only a small portion of pixel-level
annotations and a large portion of image-level annotations in the
semi-supervised setting reaps about half of this benefit.


\vspace{-0.4cm}
\paragraph{Test results}

We report our PASCAL VOC 2012 \textsl{test} results in
\tabref{tab:cross_data_test}. We include results of other
leading models from the PASCAL leaderboard. All our models have been
trained with pixel-level annotated images on the PASCAL
\textsl{trainval\_aug} and the MS-COCO 2014 \textsl{trainval}
datasets.

Methods based on the DCNN+CRF pipeline of DeepLab-CRF
\cite{chen2014semantic} are the most competitive, with performance
surpassing 70\%, even when only trained on PASCAL data. Leveraging the
MS-COCO annotations brings about 2\% improvement. Our top model
yields 73.9\%, using the multi-scale network architecture of
\cite{chen2014semantic}. Also see \cite{zheng2015conditional}, which also uses 
joint PASCAL and MS-COCO training, and further improves
performance (74.7\%) by end-to-end learning of the DCNN and CRF
parameters.

\subsection{Qualitative Segmentation Results} 
\label{sec:test_qualitative}

\begin{figure*}[th]
  \centering
  \scalebox{0.85} {
  \begin{tabular}{c c c c c c}
    \includegraphics[height=0.11\linewidth]{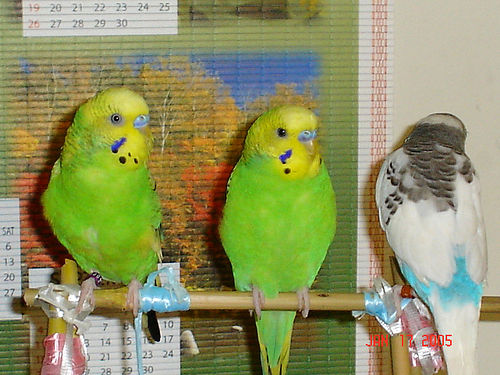} &
    \includegraphics[height=0.11\linewidth]{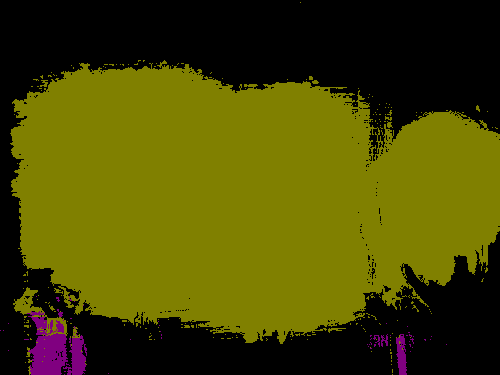} &
    \includegraphics[height=0.11\linewidth]{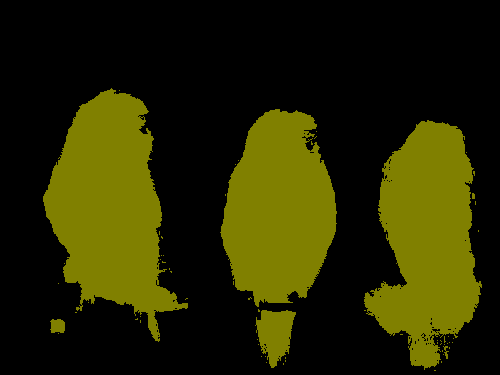} &
    \includegraphics[height=0.11\linewidth]{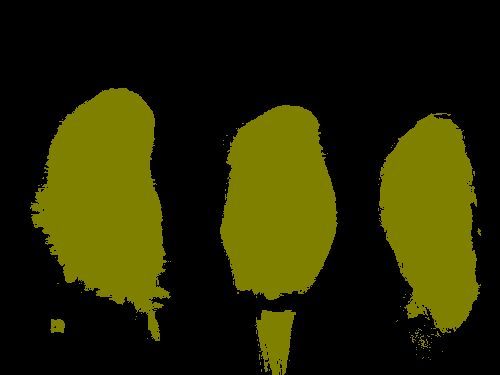} &
    \includegraphics[height=0.11\linewidth]{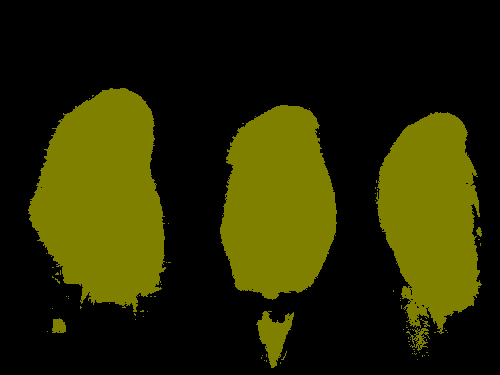} &
    \includegraphics[height=0.11\linewidth]{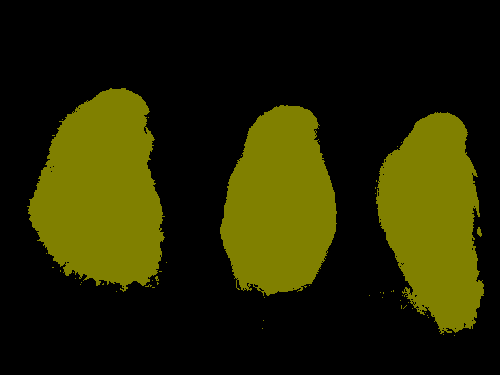} \\
    \includegraphics[height=0.11\linewidth]{fig/val_crf_vis/largewin/img/2010_000639.jpg} &
    \includegraphics[height=0.11\linewidth]{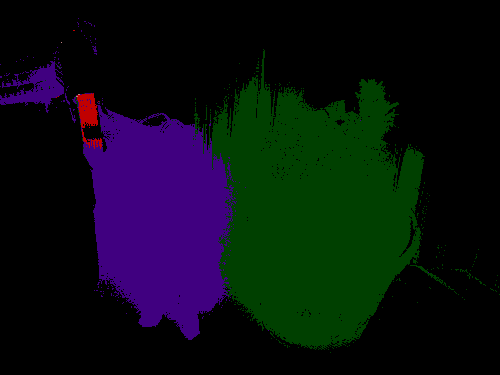} &
    \includegraphics[height=0.11\linewidth]{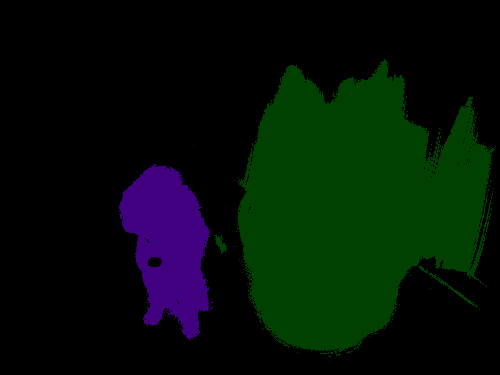} &
    \includegraphics[height=0.11\linewidth]{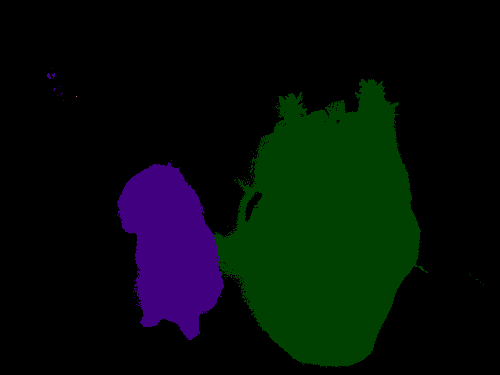} &
    \includegraphics[height=0.11\linewidth]{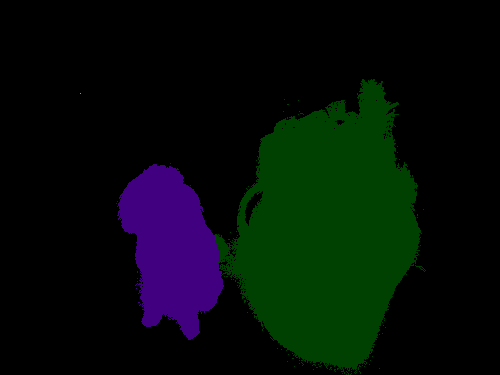} &
    \includegraphics[height=0.11\linewidth]{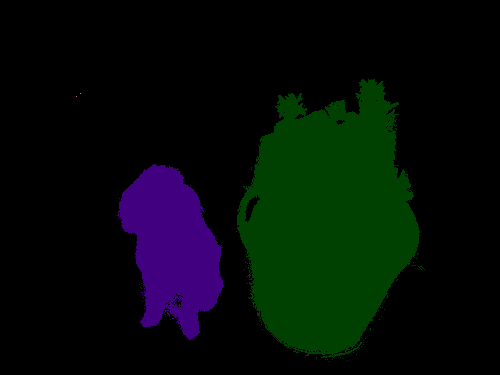} \\
    \includegraphics[height=0.098\linewidth]{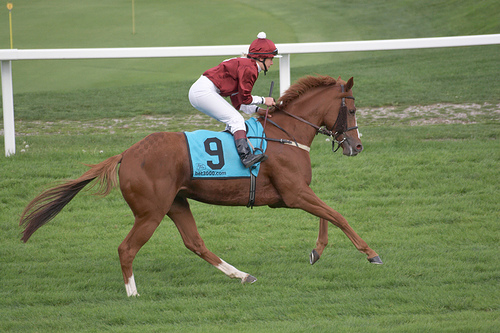} &
    \includegraphics[height=0.098\linewidth]{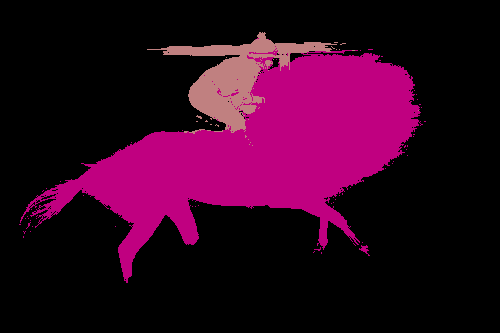} &
    \includegraphics[height=0.098\linewidth]{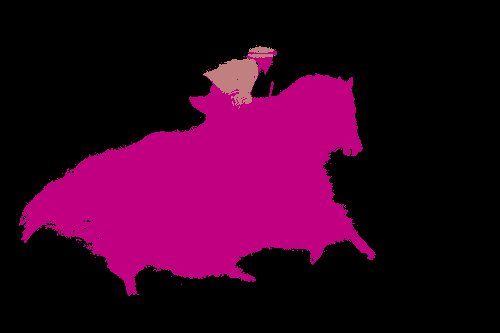} &
    \includegraphics[height=0.098\linewidth]{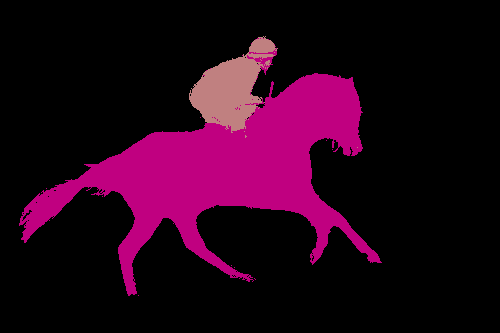} &
    \includegraphics[height=0.098\linewidth]{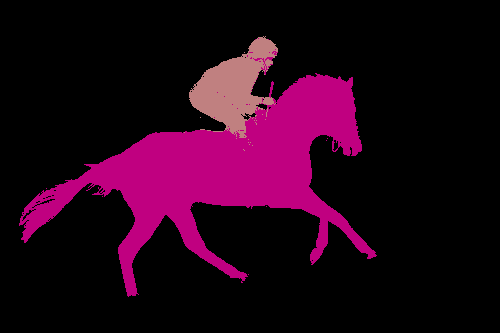} &
    \includegraphics[height=0.098\linewidth]{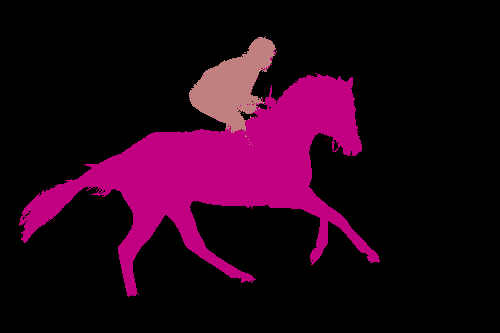} \\
    \includegraphics[height=0.11\linewidth]{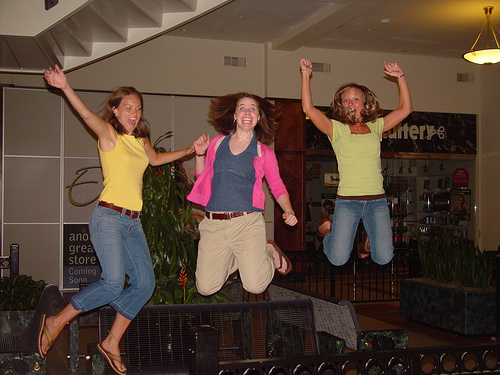} &
    \includegraphics[height=0.11\linewidth]{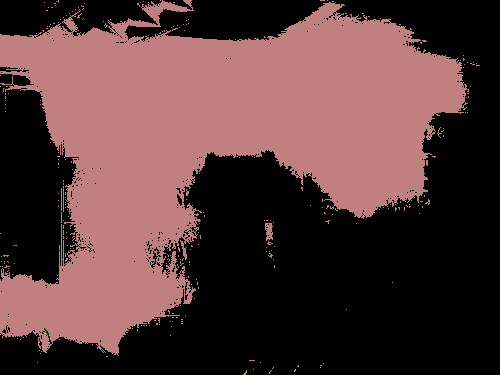} &
    \includegraphics[height=0.11\linewidth]{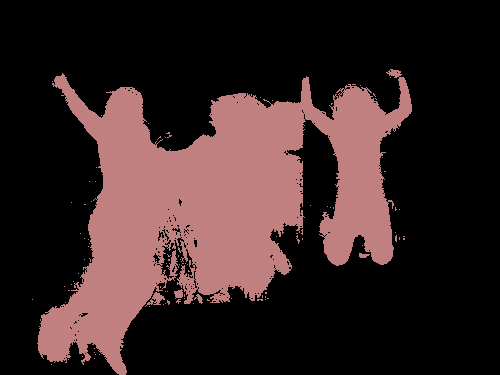} &
    \includegraphics[height=0.11\linewidth]{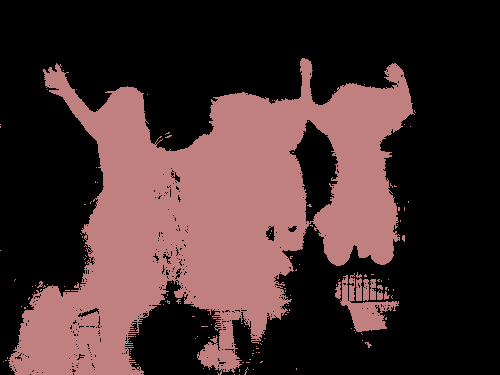} &
    \includegraphics[height=0.11\linewidth]{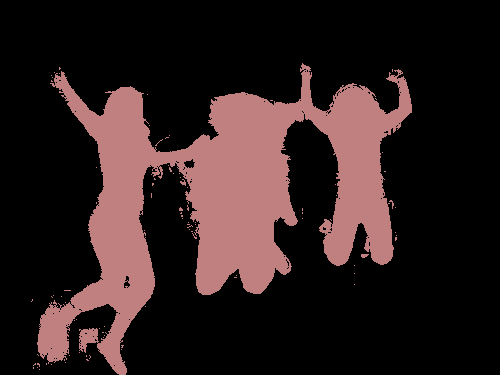} &
    \includegraphics[height=0.11\linewidth]{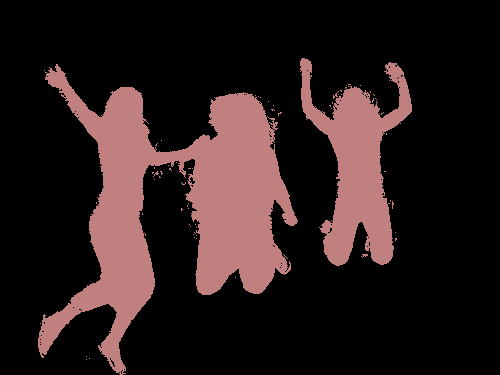} \\
    \includegraphics[height=0.11\linewidth]{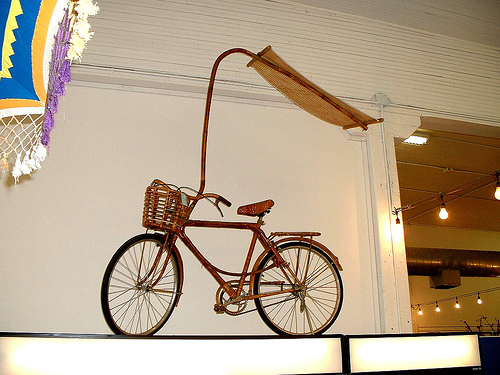} &
    \includegraphics[height=0.11\linewidth]{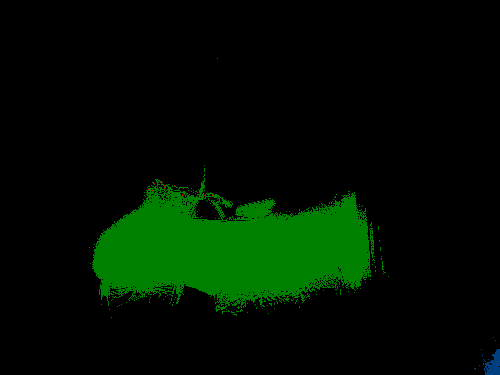} &
    \includegraphics[height=0.11\linewidth]{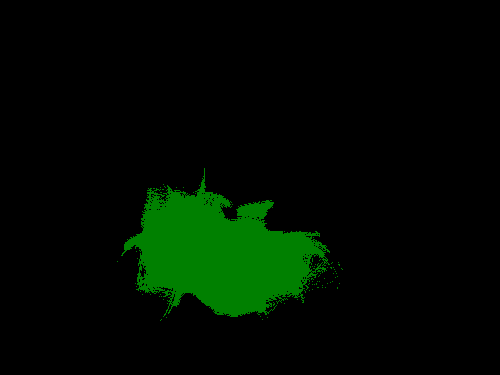} &
    \includegraphics[height=0.11\linewidth]{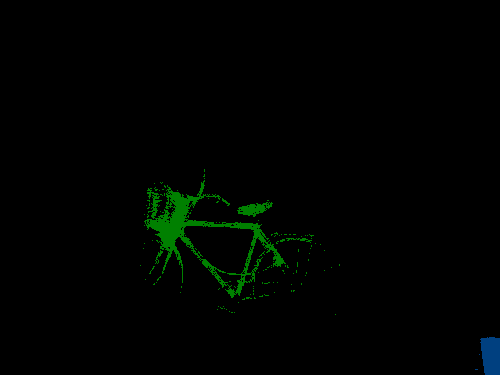} &
    \includegraphics[height=0.11\linewidth]{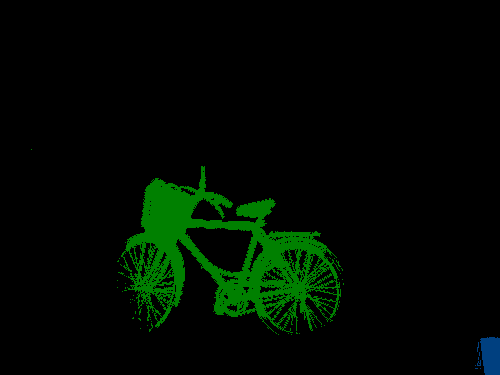} &
    \includegraphics[height=0.11\linewidth]{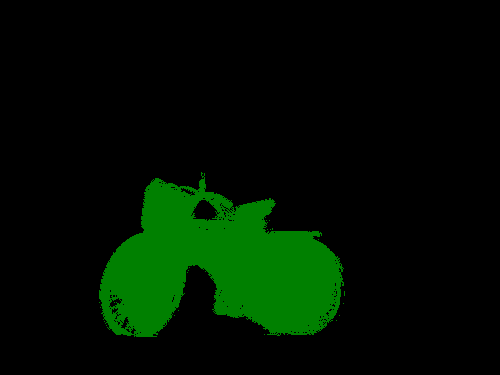} \\
    \includegraphics[height=0.11\linewidth]{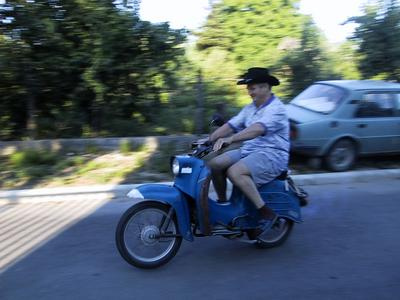} &
    \includegraphics[height=0.11\linewidth]{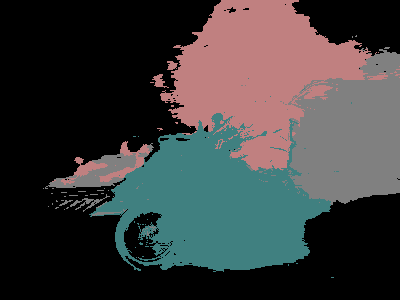} &
    \includegraphics[height=0.11\linewidth]{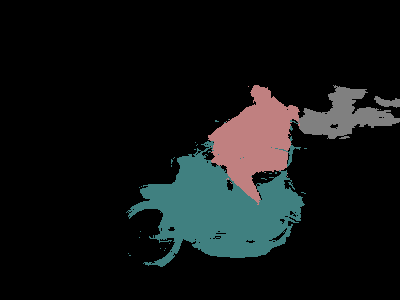} &
    \includegraphics[height=0.11\linewidth]{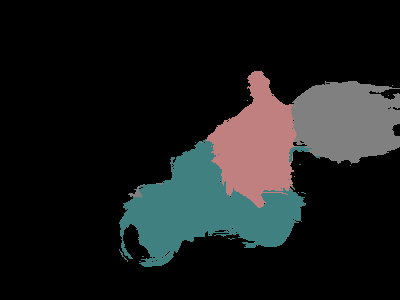} &
    \includegraphics[height=0.11\linewidth]{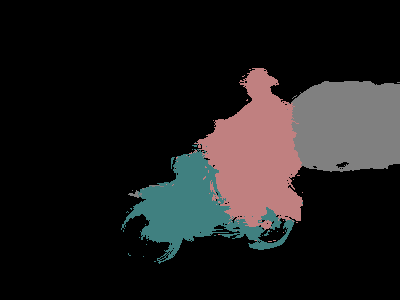} &
    \includegraphics[height=0.11\linewidth]{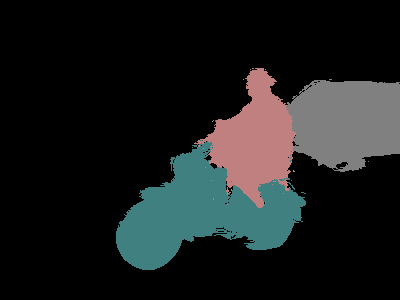} \\
    \includegraphics[height=0.11\linewidth]{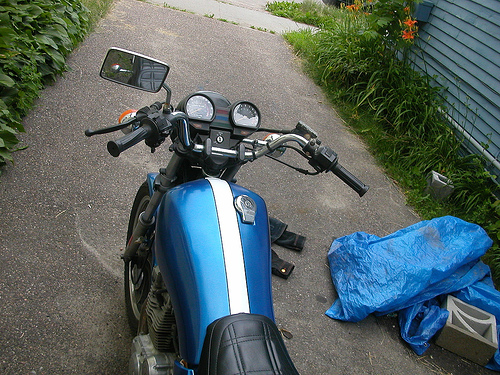} &
    \includegraphics[height=0.11\linewidth]{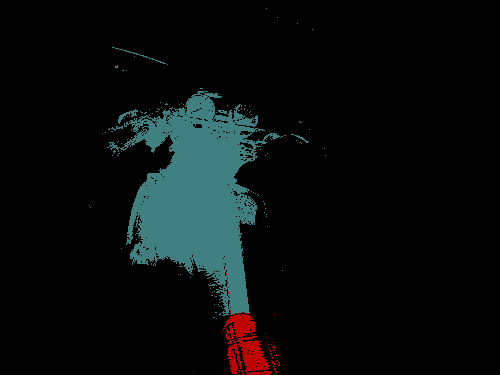} &
    \includegraphics[height=0.11\linewidth]{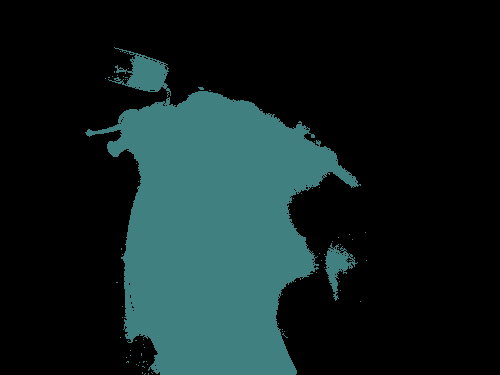} &
    \includegraphics[height=0.11\linewidth]{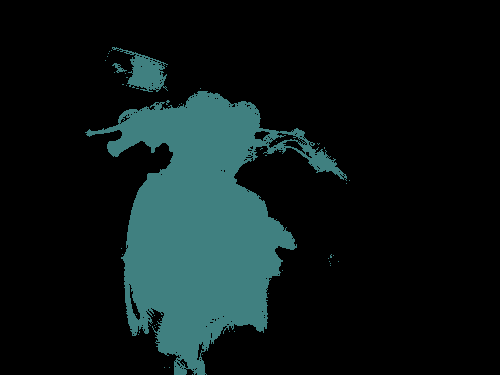} &
    \includegraphics[height=0.11\linewidth]{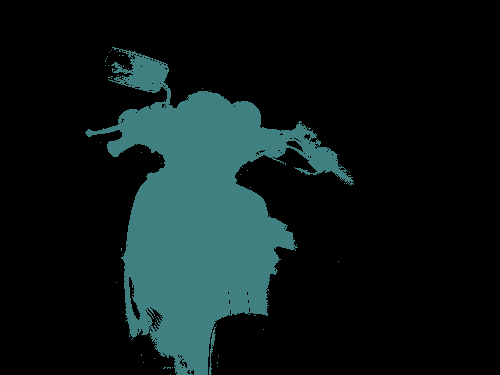} &
    \includegraphics[height=0.11\linewidth]{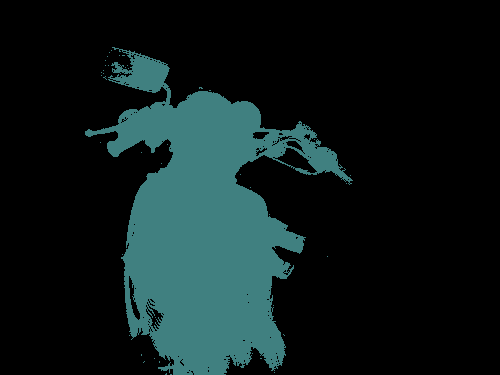} \\
    \includegraphics[height=0.11\linewidth]{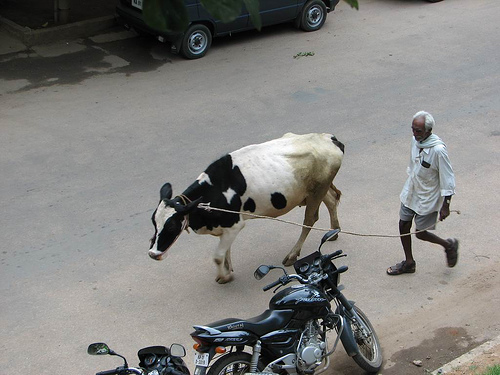} &
    \includegraphics[height=0.11\linewidth]{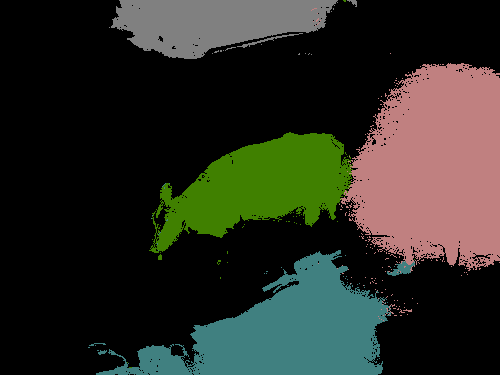} &
    \includegraphics[height=0.11\linewidth]{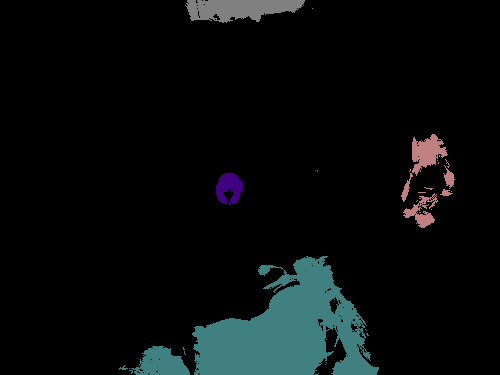} &
    \includegraphics[height=0.11\linewidth]{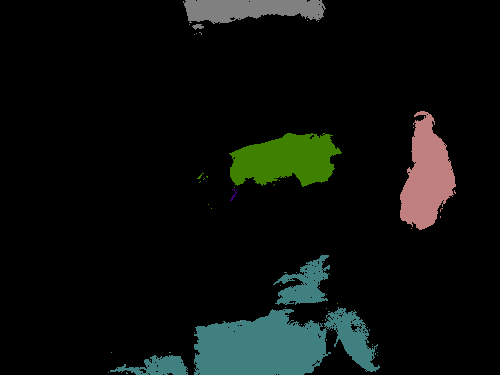} &
    \includegraphics[height=0.11\linewidth]{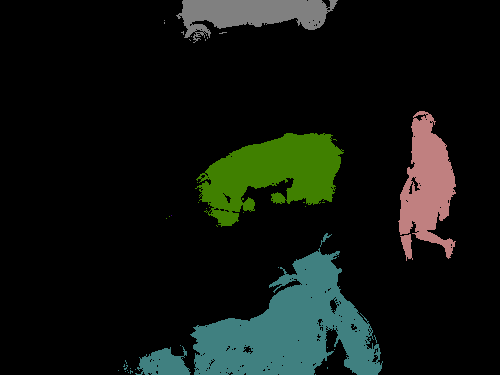} &
    \includegraphics[height=0.11\linewidth]{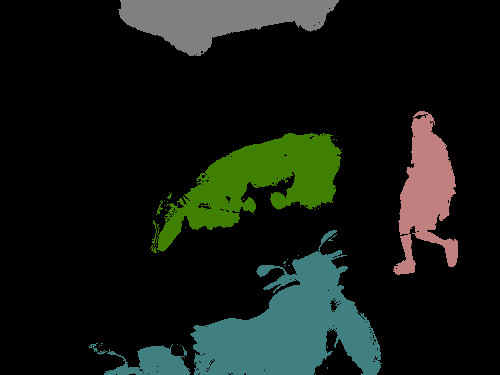} \\
    \includegraphics[height=0.11\linewidth]{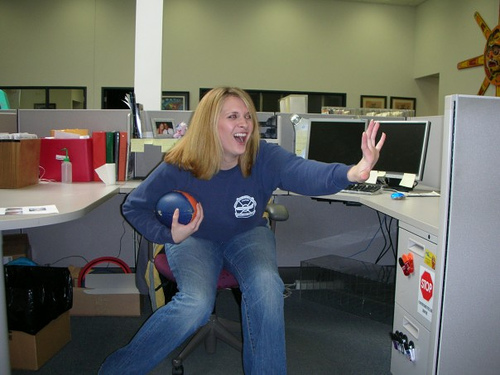} &
    \includegraphics[height=0.11\linewidth]{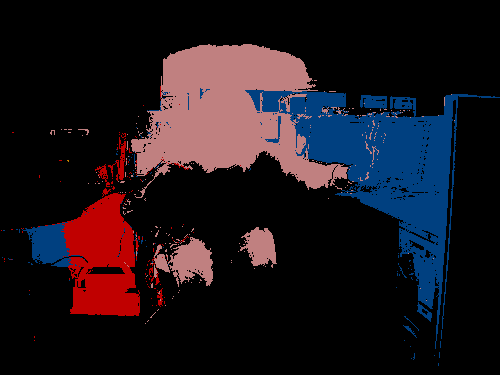} &
    \includegraphics[height=0.11\linewidth]{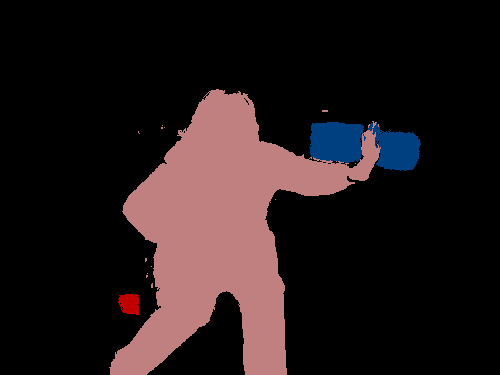} &
    \includegraphics[height=0.11\linewidth]{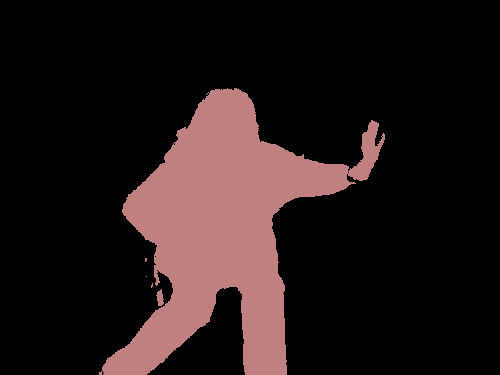} &
    \includegraphics[height=0.11\linewidth]{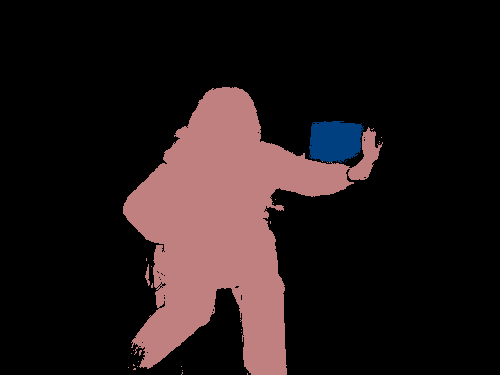} &
    \includegraphics[height=0.11\linewidth]{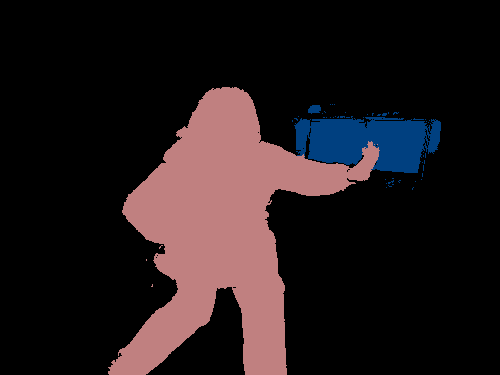} \\
    \includegraphics[height=0.15\linewidth]{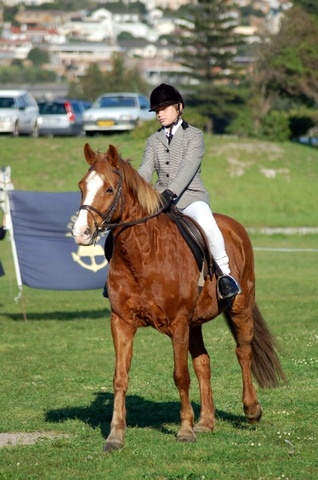} &
    \includegraphics[height=0.15\linewidth]{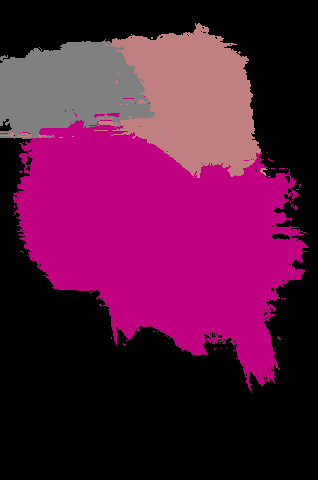} &
    \includegraphics[height=0.15\linewidth]{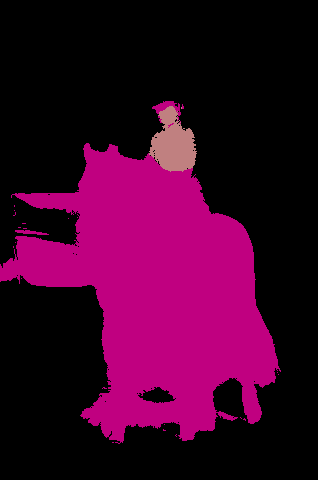} &
    \includegraphics[height=0.15\linewidth]{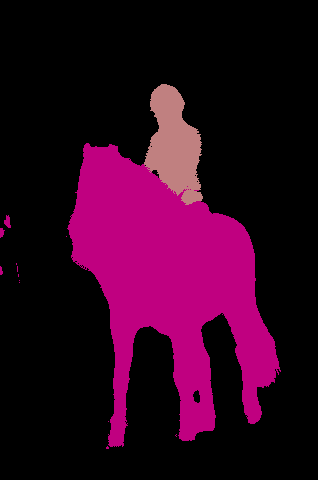} &
    \includegraphics[height=0.15\linewidth]{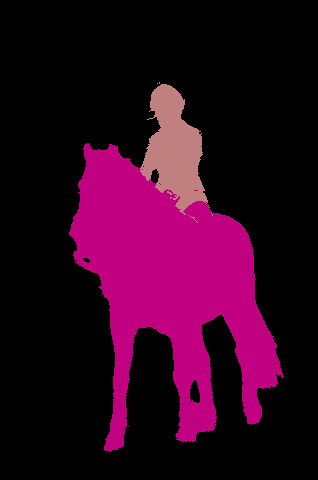} &
    \includegraphics[height=0.15\linewidth]{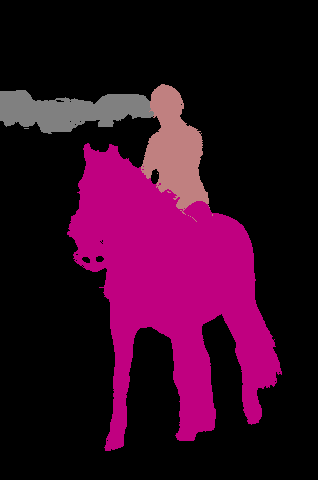} \\
    \hline \hline
    \includegraphics[height=0.11\linewidth]{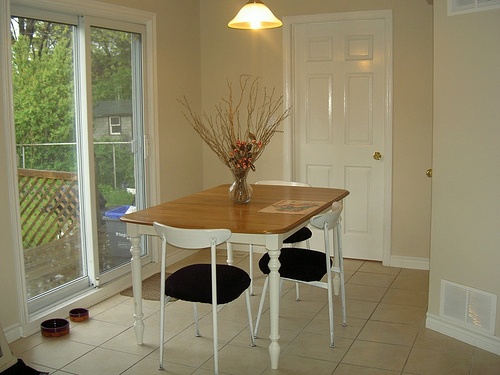} &
    \includegraphics[height=0.11\linewidth]{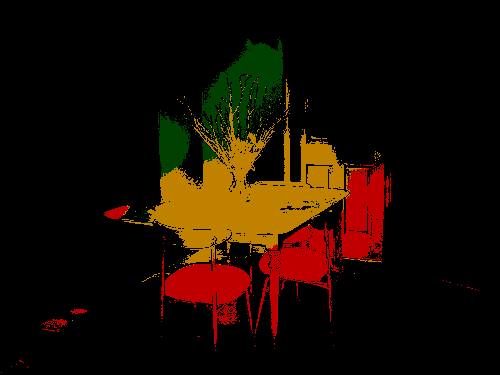} &
    \includegraphics[height=0.11\linewidth]{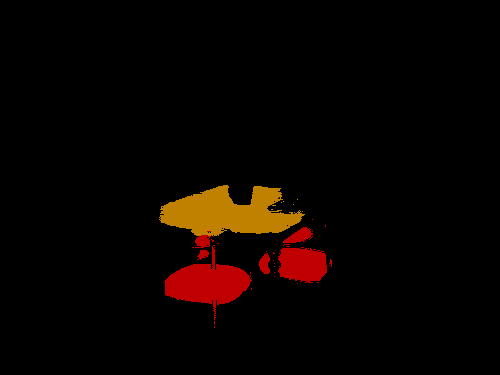} &
    \includegraphics[height=0.11\linewidth]{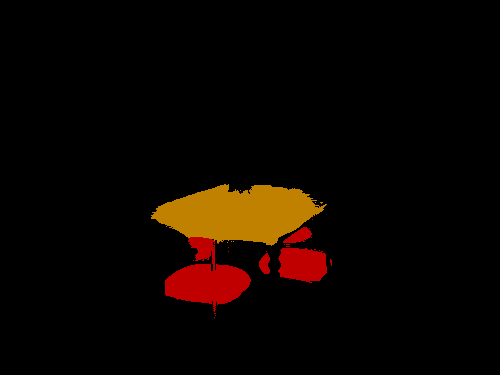} &
    \includegraphics[height=0.11\linewidth]{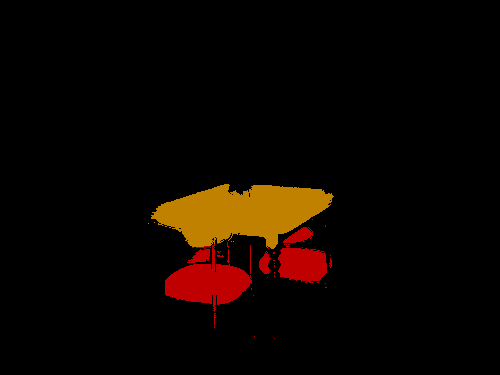} &
    \includegraphics[height=0.11\linewidth]{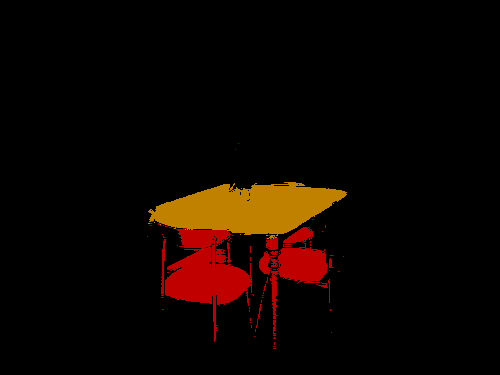} \\
    \includegraphics[height=0.11\linewidth]{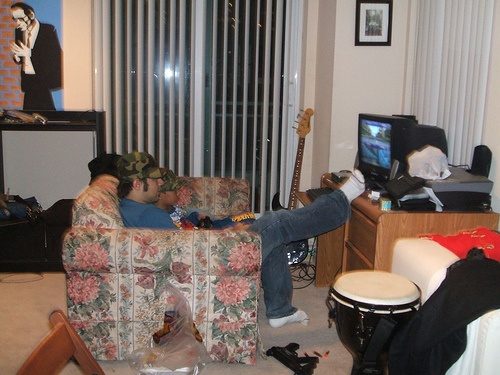} &
    \includegraphics[height=0.11\linewidth]{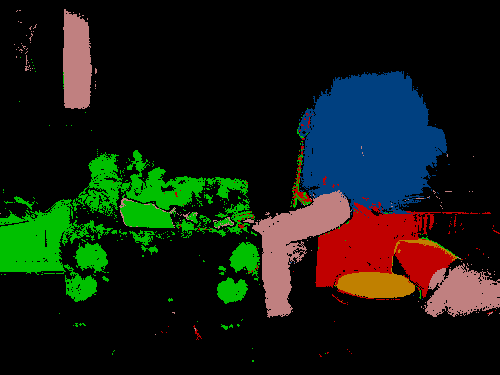} &
    \includegraphics[height=0.11\linewidth]{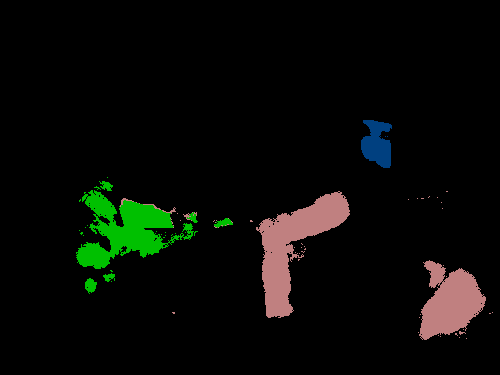} &
    \includegraphics[height=0.11\linewidth]{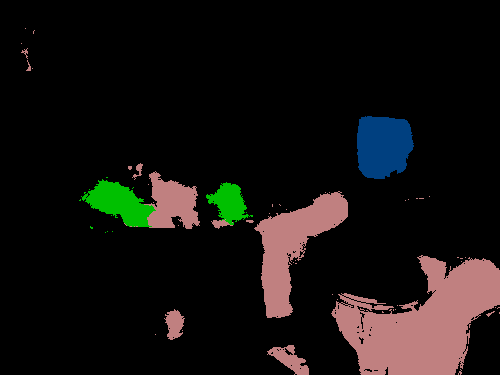} &
    \includegraphics[height=0.11\linewidth]{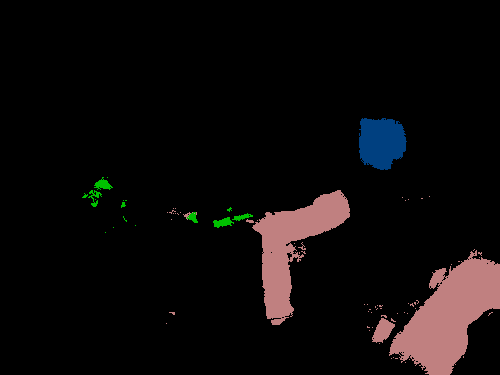} &
    \includegraphics[height=0.11\linewidth]{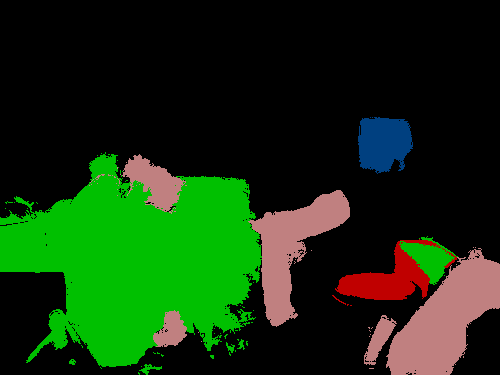} \\

    {\footnotesize Image} & {\footnotesize EM-Adapt (Weak)} & {\footnotesize
      Bbox-Seg (Weak)} & {\footnotesize EM-Fixed (Semi)} & {\footnotesize
      Bbox-EM-Fixed (Semi)} & {\footnotesize Cross-Joint (Strong)} \\
  \end{tabular}
  }
  \caption{Qualitative DeepLab-CRF segmentation results on the PASCAL
    VOC 2012 \textsl{val} set. The last two rows show failure modes.}
  \label{fig:ValResults}
\end{figure*}

In \figref{fig:ValResults} we provide visual comparisons of the
results obtained by the DeepLab-CRF model learned with some of the
proposed training methods.



\section{Conclusions}

The paper has explored the use of weak or partial annotation
in training a state of art semantic image segmentation
model. Extensive experiments on the challenging PASCAL VOC 2012
dataset have shown that: (1) Using weak annotation solely at the
image-level seems insufficient to train a high-quality segmentation
model. (2) Using weak bounding-box annotation in conjunction with
careful segmentation inference for images in the training set suffices
to train a competitive model. (3) Excellent performance is obtained when
combining a small number of pixel-level annotated images with a large
number of weakly annotated images in a semi-supervised setting,
nearly matching the results achieved when all training images have
pixel-level annotations. (4) Exploiting extra weak or strong
annotations from other datasets can lead to large improvements.

\subsection*{Acknowledgments} 
This work was partly supported by ARO 62250-CS, and NIH 5R01EY022247-03. 
We also gratefully acknowledge the support of NVIDIA Corporation with the 
donation of GPUs used for this research.



\appendix

\section*{Supplementary Material}

We include as appendix: (1) Details of the proposed EM-Adapt algorithm. (2) More experimental evaluations about the effect of the model's Field-Of-View. (3) More detailed results of the proposed training methods on PASCAL VOC 2012 test set.

\section{E-Step with Cardinality Constraints: Details of our EM-Adapt Algorithm}

Herein we provide more background and a detailed description of our
\textsl{EM-Adapt} algorithm for weakly-supervised training with
image-level annotations.

As a reminder, $\yv$ is the latent segmentation map,
with $y_m \in \{0, \ldots, L\}$ denoting the label at position $m \in
\{1,\ldots,M\}$. The image-level annotation is encoded in  $\zv$, with
$z_l = 1$, if the $l$-th label is present anywhere in the image.

We assume that $\log P(\zv | \yv) = \phi(\yv, \zv) +
\mathrm{(const)}$. We employ a cardinality potential $\phi(\yv, \zv)$
which encourages at least a $\rho_l$ portion of the image area to be
assigned to class $l$, if $z_l = 1$, and enforce that no pixel is
assigned to class $l$, if $z_l = 0$. We set the parameters $\rho_l =
\rho_{\mathrm{fg}}$, if $l > 0$ and $\rho_0 = \rho_{\mathrm{bg}}$.

While dedicated algorithms exist for optimizing energy functions under
such cardinality potentials \cite{tarlow2012fast,
  pletscher2012learning, Li2014high}, we opt for a simpler alternative
that approximately enforces these area constraints and works well in
practice. We use a variant of the \textsl{EM-Fixed} algorithm described
in the main paper, updating the segmentations in the E-Step by
$\hat{y}_m = \argmax_{l} \hat{f}_m (l) \doteq f_m(l | \xv; \thetav') +
b_l$. The key difference in the \textsl{EM-Adapt} variant is that the
biases $b_l$ are \emph{adaptively} set so as the prescribed proportion
of the image area is assigned to the background or foreground object
classes that are present in the image.

When only one label $l$ is present (\ie $z_l = 1$, $\sum_{l'=0}^L
z_{l'} = 1$), one can easily enforce the constraint that at least
$\rho_l$ of the image area is assigned to label $l$ as follows:
(1) Set $b_{l'} = 0, l' \neq l$. (2) Compute the maximum score at each
position, $f_m^{\mathrm{max}} = \max_{l'=0}^L f_m(l' | \xv;
\thetav')$. (3) Set $b_l$ equal to the $\rho_l$-th
percentile of the score difference $d_m = f_m^{\mathrm{max}} - f_m(l |
\xv; \thetav')$. The cost of this algorithm is $\mathcal{O}(M)$
(linear \wrt the number of pixels).

When more than one labels are present in the image (\ie $\sum_{l'=0}^L
z_{l'} > 1$), we employ the procedure above sequentially for each
label that $z_l > 1$ (we first visit the background label, then in
random order each of the foreground labels which are present in the
image). We set $b_l = -\infty$, if $z_l = 0$, to suppress the labels
that are not present in the image.

An implementation of this algorithm will become publicly available
after this paper gets published.

\section{Effect of Field-Of-View}
In this section, we explore the effect of Field-Of-View (FOV) when training the DeepLab-CRF model with the proposed methods in the main paper. Similar to \cite{chen2014semantic}, we also employ the `atrous' algorithm \cite{Mall99} in the DeepLab model. The `atrous' algorithm enables us to arbitrary control the model's FOV by adjusting the input stride (which is equivalent to injecting zeros between filter weights) at the first fully connected layer of VGG-16 net \cite{simonyan2014very}. Applying a large value of input stride increases the effective kernel size, and thus enlarges the model's FOV (see \cite{chen2014semantic} for details).

\paragraph{Experimental protocol} We employ the same experimental protocol as the main paper. Models trained with the proposed training methods and different values of FOV are evaluated on the PASCAL VOC 2012 \textsl{val} set.

\paragraph{EM-Adapt} Assuming only image-level labels are available, we first experiment with the \textsl{EM-Adapt (Weak)} method when the value of FOV varies. Specifically, we explore the setting where the kernel size is \by{3}{3} with various FOV values. The selection of kernel size \by{3}{3} is based on the discovery by \cite{chen2014semantic}: employing a kernel size of \by{3}{3} at the first fully connected layer can attain the same performance as using the kernel size of \by{7}{7}, while being $3.4$ times faster during training. As shown in \tabref{tab:sup_fov}, we find that our proposed model can yield the performance of $39.2\%$ with FOV \by{96}{96}, but the performance degrades by $9\%$ when large FOV \by{224}{224} is employed. The original DeepLab model employed by \cite{chen2014semantic} has a kernel size of \by{4}{4} and input stride of 4. Its performance, shown in the first row of \tabref{tab:sup_fov}, is similar to the performance obtained by using a kernel size of \by{3}{3} and input stride of 6. Both cases have the same FOV value of \by{128}{128}.

\paragraph{Network architectures} In the following experiments, we compare two network architectures trained with the methods proposed in the main paper. The first network is the same as the one originally employed by \cite{chen2014semantic} (kernel size \by{4}{4} and input stride 4, resulting in a FOV size \by{128}{128}). The second network we employ has FOV \by{224}{224} (with kernel size of \by{3}{3} and an input stride of 12). We refer to the first network as `DeepLab-CRF with small FOV', and the second network as `DeepLab-CRF with large FOV'.

\paragraph{Image-level annotations} In \tabref{tab:sup_image_annot}, we experiment with the cases where weak image-level annotations as well as a varying number of pixel-level annotations are available. Similar to the results in \tabref{tab:sup_fov}, the DeepLab-CRF with small FOV performs better than that with large FOV when a small amount of supervision is leveraged. Interestingly, when there are more than $750$ pixel-level annotations are available in the semi-supervised setting, employing large FOV yields better performance than using small FOV.

\paragraph{Bounding box annotations} In \tabref{tab:sup_bbox_annot}, we report the results when weak bounding box annotations in addition to a varying number of pixel-level annotations are exploited. we found that DeepLab-CRF with small FOV attains better performance when trained with the three methods: Bbox-Rect (Weak), Bbox-EM-Fixed (Weak), and Bbox-Rect (Semi- 1464 strong), whereas the model DeepLab-CRF with large FOV is better in all the other cases.

\paragraph{Annotations across datasets} In \tabref{tab:sup_cross_data}, we show the results when training the models with the strong pixel-level annotations from PASCAL VOC 2012 \textsl{train\_aug} set in conjunction with the extra annotations from MS-COCO \cite{lin2014microsoft} dataset (in the form of either weak image-level annotations or strong pixel-level annotations). Interestingly, employing large FOV consistently improves over using small FOV in all cases by at least 3\%.

\paragraph{Main paper} Note that in the main paper we report the results of the best architecture for each setup.

\section{Detailed test results}
In \tabref{tab:sup_image_annot_test}, \tabref{tab:sup_bbox_annot_test}, and \tabref{tab:sup_cross_data_test}, we show more detailed results on PASCAL VOC 2012 \textsl{test} set for all the reported methods in the main paper.

\begin{table}[tbp!]
  \centering
  \scalebox{0.7}{
  \begin{tabular}{|c | c | c | c |}
    \hline
    kernel size & input stride & receptive field & val IOU (\%) \\
    \hline 
    \by{4}{4}         &    4  & \by{128}{128} & 38.2 \\
    \hline \hline
    \by{3}{3}         &    2  & \by{64}{64} &  37.3 \\
    \hline
    \by{3}{3}         &    4  & \by{96}{96} &  39.2 \\
    \hline
    \by{3}{3}         &    6  & \by{128}{128} &  38.3 \\
    \hline
    \by{3}{3}         &    8  & \by{160}{160} &  38.1 \\
    \hline
    \by{3}{3}         &    10  & \by{192}{192} &  32.6 \\
    \hline
    \by{3}{3}         &    12  & \by{224}{224} &  30.2 \\
    \hline
  \end{tabular}
  }
  \caption{Effect of Field-Of-View. The validation performance obtained by DeepLab-CRF trained with the method EM-Adapt (Weak) as the value of FOV varies.}
  \label{tab:sup_fov}
\end{table}


\begin{table}[tbp!]
  \centering
  \scalebox{0.7}{
  \begin{tabular}{|c|c|c|c|c|}
    \hline
    Method  & \#Strong & \#Weak  & w Small FOV &  w Large FOV \\
    \hline
    EM-Fixed (Weak)                   &  -       &  10,582 & 20.8     &  19.9   \\
    EM-Adapt (Weak)                   &  -       &  10,582 & 38.2     &  30.2   \\
    \hline
    \multirow{6}{*}{EM-Fixed (Semi)}  & 200      &  10,382 & 47.6     &  38.9    \\
                                      & 500      &  10,082 & 56.9     &  54.2  \\
                                      & 750      &  9,832  & 58.8     &  59.8  \\
                                      & 1,000    &  9,582  & 60.5     &  62.0  \\
                                      & 1,464    &  5,000  & 60.5     &  63.2   \\
                                      & 1,464    &  9,118  & 61.9     &  64.6  \\
    \hline
     \multirow{2}{*}{Strong}          & 1,464    &  -      & 57.6     &  62.5  \\
                                      & 10,582   &  -      & 63.9     &  67.6 \\
    \hline
  \end{tabular}  
  }
  \caption{Effect of Field-Of-View. VOC 2012 \textsl{val} performance for varying number of pixel-level (strong) and image-level (weak) annotations
    (Sec.~4.3 of main paper).}
  \label{tab:sup_image_annot}
\end{table}

\begin{table}[tbp!]
  \centering
  \scalebox{0.7}{
  \begin{tabular}{|c|c|c|c|c|}
    \hline
    Method & \#Strong & \#Box  & w Small FOV & w Large FOV  \\
    \hline
    Bbox-Rect (Weak)        &   -        &  10,582 & 52.5  & 50.7 \\
    Bbox-EM-Fixed (Weak)    &   -        &  10,582 & 54.1  & 50.2 \\
    Bbox-Seg (Weak)         &   -        &  10,582 & 58.5  & 60.6 \\
    \hline 
    Bbox-Rect (Semi)        &   1,464    &  9.118  & 62.1  & 61.1 \\
    Bbox-EM-Fixed (Semi)    &   1,464    &  9,118  & 59.6  & 64.8 \\
    Bbox-Seg (Semi)         &   1,464    &  9,118  & 61.8  & 65.1 \\
    \hline 
    Strong                  &   1,464    &  -      & 57.6  & 62.5 \\
    Strong                  &   10,582   &  -      & 63.9  & 67.6 \\
    \hline
  \end{tabular}  
  }
  \caption{Effect of Field-Of-View. VOC 2012 \textsl{val} performance for varying number of pixel-level (strong) and bounding box (weak) annotations
    (Sec.~4.4 of main paper).}
  \label{tab:sup_bbox_annot}
\end{table}

\begin{table}[tbp!]
  \centering
  \scalebox{0.7}{
  \begin{tabular}{|c|c|c|c|c|}
    \hline
    Method & \#Strong & \#Weak  & w Small FOV & w Large FOV \\
    \hline
    PASCAL-only             & -        & -       & 63.9        &  67.6  \\
    \hline
    EM-Fixed (Semi)         & -        & 123,287 & 64.4        &  67.7  \\
    \hline
    Cross-Joint (Semi)      & 5,000    & 118,287 & 66.5        &  70.0   \\
    \hline
    Cross-Joint (Strong)    & 5,000    &  -      & 64.9        &  68.7  \\
    Cross-Pretrain (Strong) & 123,287  &  -      & 68.0        &  71.0  \\
    Cross-Joint (Strong)    & 123,287  &  -      & 68.0        &  71.7  \\
    \hline
  \end{tabular}  
  }
  \caption{Effect of Field-Of-View. VOC 2012 \textsl{val} performance using strong annotations for all 10,582 \textsl{train\_aug} PASCAL images and a varying
    number of strong and weak MS-COCO annotations
    (Sec.~4.5 of main paper).}
  \label{tab:sup_cross_data}
\end{table}

\begin{table*}[t!]\scriptsize
\setlength{\tabcolsep}{3pt}
\resizebox{2.1\columnwidth}{!}{
\begin{tabular}{|c||c*{20}{|c}||c|}
\hline 
Method          & bkg &  aero & bike & bird & boat & bottle& bus & car  &  cat & chair& cow  &table & dog  & horse & mbike& person& plant&sheep& sofa &train & tv   & mean \\
\hline \hline
MIL-FCN \cite{pathak2014fully} & - & - & -& -& -& -& -& -& -& -& -& -& -& -& -& -& -& -& -& -& - & 25.7 \\
MIL-sppxl \cite{Pinheiro2015}  & 74.7 & 38.8 & 19.8 & 27.5 & 21.7 & 32.8 & 40.0 & 50.1 & 47.1 & 7.2 & 44.8 & 15.8 & 49.4 & 47.3 & 36.6 & 36.4 & 24.3 & 44.5 & 21.0 & 31.5 & 41.3 & 35.8 \\
MIL-obj \cite{Pinheiro2015} & 76.2 & 42.8 & 20.9 & 29.6 & 25.9 & 38.5 & 40.6 & 51.7 & 49.0 & 9.1 & 43.5 & 16.2 & 50.1 & 46.0 & 35.8 & 38.0 & 22.1 & 44.5 & 22.4 & 30.8 & 43.0 & 37.0 \\
MIL-seg \cite{Pinheiro2015}        & 78.7 & 48.0 & 21.2 & 31.1 & 28.4 & 35.1 & 51.4 & 55.5 & 52.8 & 7.8 & 56.2 & 19.9 & 53.8 & 50.3 & 40.0 & 38.6 & 27.8 & 51.8 & 24.7 & 33.3 & 46.3 & 40.6 \\
\hline \hline
\href{http://host.robots.ox.ac.uk:8080/anonymous/U0IL1U.html}{EM-Adapt (Weak)} & 76.3 & 37.1 & 21.9 & 41.6 & 26.1 & 38.5 & 50.8 & 44.9 & 48.9& 16.7 & 40.8 & 29.4 & 47.1 & 45.8 & 54.8 & 28.2 & 30.0 & 44.0 & 29.2 & 34.3 & 46.0 &
39.6 \\
\hline
\href{http://host.robots.ox.ac.uk:8080/anonymous/XDC2GJ.html}{EM-Fixed (Semi-1464 strong)} & 91.3 & 78.9 & 37.3 & {\bf 81.4} & 57.1 &  57.7 & 83.5 & 77.5 & {\bf 77.6} & 22.5 & 70.3 & 56.1 & 72.2 & {\bf 74.3} & 80.7 & 72.4 & 42.0 & 81.3 & 43.1 & 72.5 & 60.7 & 66.2 \\
\hline
\href{http://host.robots.ox.ac.uk:8080/anonymous/BUUC2K.html}{EM-Fixed (Semi-2913 strong)} & {\bf 92.0} & {\bf 81.6} & {\bf 42.9} & 80.5 & {\bf 59.2} & {\bf 60.8} & {\bf 85.5} & {\bf 78.7} &  77.3 & {\bf 26.9} & {\bf 75.2} & {\bf 57.6} & {\bf 74.0} & 74.2 & {\bf 82.1} & {\bf 73.1} & {\bf 52.4} & {\bf 84.3} & {\bf 43.8} & {\bf 75.1} & {\bf 61.9} & {\bf 68.5} \\
\hline
 \end{tabular}
} 
 \caption{VOC 2012 \textsl{test} performance for varying number of pixel-level (strong) and image-level (weak) annotations (Sec.~4.3 of main paper). Links to the PASCAL evaluation server are included in the PDF.}
 \label{tab:sup_image_annot_test}
\end{table*}

\begin{table*}[t!]\scriptsize
\setlength{\tabcolsep}{3pt}
\resizebox{2.1\columnwidth}{!}{
\begin{tabular}{|c||c*{20}{|c}||c|}
\hline 
Method          & bkg &  aero & bike & bird & boat & bottle& bus & car  &  cat & chair& cow  &table & dog  & horse & mbike& person& plant&sheep& sofa &train & tv   & mean \\
\hline \hline
BoxSup-box \cite{dai2015boxsup} & - & 80.3 & 31.3 & 82.1 & 47.4 & 62.6 & 75.4 & 75.0 & 74.5 & 24.5 & 68.3 & 56.4 & 73.7 & 69.4 & 75.2 & 75.1 & 47.4 & 70.8 & 45.7 & 71.1 & 58.8 & 64.6 \\
BoxSup-semi \cite{dai2015boxsup} & - & {\bf 82.0} & 33.6 & 74.0 & 55.8 & 57.5 & 81.0 & 74.6 & 80.7 & 27.6 & 70.9 & 50.4 & 71.6 & 70.8 & 78.2 & 76.9 & 53.5 & 72.6 & {\bf 50.1} & 72.3 & {\bf 64.4} & 66.2 \\
\hline \hline
\href{http://host.robots.ox.ac.uk:8080/anonymous/UKB05Q.html}{Bbox-Rect (Weak)}
                & 82.9 & 43.6 & 22.5 & 50.5 & 45.0 & 62.5 & 76.0 & 66.5 & 61.2 & 25.3 & 55.8 & 52.1 & 56.6 & 48.1 & 60.1 & 58.2 & 49.5 & 58.3 & 40.7 & 62.3 & 61.1 & 54.2 \\
\href{http://host.robots.ox.ac.uk:8080/anonymous/BC5JKR.html}{Bbox-Seg (Weak)} & 89.2 & 64.4 & 27.3 & 67.6 & 55.1 & 64.0 & 81.6 & 70.5 & 76.0& 24.1 & 63.8 & {\bf 58.2} & 72.1 & 59.8 & 73.5 & 71.4 & 47.4 & 76.0 & 44.2 & 68.9 & 50.9 & 62.2 \\
\hline
\href{http://host.robots.ox.ac.uk:8080/anonymous/VE7S2X.html}{Bbox-Seg (Semi-1464 strong)} & 91.3 & 75.3 & 29.9 & 74.4 & 59.8 & 64.6 & 84.3 & 76.2 & 79.0 & 27.9 & 69.1 & 56.5 &  73.8 & 66.7 & 78.8 & 76.0 & 51.8 & 80.8 & 47.5 & 73.6 & 60.5 & 66.6  \\
\href{http://host.robots.ox.ac.uk:8080/anonymous/KM7AMP.html}{Bbox-EM-Fixed (Semi-1464 strong)} & 91.9 & 78.3 & 36.5 & {\bf 86.2} & 53.8 & 62.5 & 81.2 & {\bf 80.0} & 83.2 & 22.8 & 68.9 & 46.7 & 78.1 & 72.0 & 82.2 & 78.5 & 44.5 & 81.1 & 36.4 & 74.6 & 60.2 & 66.6 \\
\hline
\href{http://host.robots.ox.ac.uk:8080/anonymous/O04D27.html}{Bbox-Seg (Semi-2913 strong)} & 92.0 & 76.4 & 34.1 & 79.2 & {\bf 61.0} & {\bf 65.6} & {\bf 85.0} & 76.9 & 81.5 & {\bf 28.5} & 69.3 & 58.0 & 75.5 & 69.8 & 79.3 & 76.9 & {\bf 54.0} & 81.4 & 46.9 & 73.6 & 62.9 & 68.0 \\
\href{http://host.robots.ox.ac.uk:8080/anonymous/R40ESI.html}{Bbox-EM-Fixed (Semi-2913 strong)} & {\bf 92.5} & 80.4 & {\bf 41.6} & 84.6 & 59.0 & 64.7 & 84.6 & 79.6 & {\bf 83.5} & 26.3 & {\bf 71.2} & 52.9 & {\bf 78.3} & {\bf 72.3} & {\bf 83.3} & {\bf 79.1} & 51.7 & {\bf 82.1} & 42.5 & {\bf 75.0} & 63.4 & {\bf 69.0} \\
\hline
 \end{tabular}
} 
 \caption{VOC 2012 \textsl{test} performance for varying number of
   pixel-level (strong) and bounding box (weak) annotations (Sec.~4.4 of main paper). Links to the PASCAL evaluation server are included in the PDF.}
 \label{tab:sup_bbox_annot_test}
\end{table*}

\begin{table*}[t!]\scriptsize
\setlength{\tabcolsep}{3pt}
\resizebox{2.1\columnwidth}{!}{
\begin{tabular}{|c||c*{20}{|c}||c|}
\hline 
Method          & bkg &  aero & bike & bird & boat & bottle& bus & car  &  cat & chair& cow  &table & dog  & horse & mbike& person& plant&sheep& sofa &train & tv   & mean \\
\hline \hline
MSRA-CFM \cite{dai2014convolutional}       & -    & 75.7 & 26.7 & 69.5 & 48.8 & 65.6 & 81.0 & 69.2 & 73.3 & 30.0 & 68.7 & 51.5 & 69.1 & 68.1 & 71.7 & 67.5 & 50.4 & 66.5 & 44.4 & 58.9 & 53.5 & 61.8 \\
FCN-8s \cite{long2014fully}        & -    & 76.8 & 34.2 & 68.9 & 49.4 & 60.3 & 75.3 & 74.7 & 77.6 & 21.4 & 62.5 & 46.8 & 71.8 & 63.9 & 76.5 & 73.9 & 45.2 & 72.4 & 37.4 & 70.9 & 55.1 & 62.2 \\
Hypercolumn \cite{hariharan2014hypercolumns} & - & 68.7 & 33.5 & 69.8 & 51.3 & {\bf 70.2} & 81.1 & 71.9 & 74.9 & 23.9 & 60.6 & 46.9 & 72.1 & 68.3 & 74.5 & 72.9 & 52.6 & 64.4 & 45.4 & 64.9 & 57.4 & 62.6 \\   
TTI-Zoomout-16 \cite{mostajabi2014feedforward}  & 89.8 & 81.9 & 35.1 & 78.2 & 57.4 & 56.5 & 80.5 & 74.0 & 79.8 & 22.4 & 69.6 & 53.7 & 74.0 & 76.0 & 76.6 & 68.8 & 44.3 & 70.2 & 40.2 & 68.9 & 55.3 & 64.4 \\
CRF\_RNN \cite{zheng2015conditional} & - & 80.9 & 34.0 & 72.9 & 52.6 & 62.5 & 79.8 & 76.3 & 79.9 & 23.6 & 67.7 & 51.8 & 74.8 & 69.9 & 76.9 & 76.9 & 49.0 & 74.7 & 42.7 & 72.1 & 59.6 & 65.2 \\
DeepLab-CRF-LargeFOV \cite{chen2014semantic} & 92.6 & 83.5 & 36.6 & 82.5 & 62.3 & 66.5 & 85.4 & 78.5 & 83.7 & 30.4 & 72.9 & 60.4 & 78.5 & 75.5 & 82.1 & 79.7 &  58.2 & 82.0 & 48.8 & 73.7 & 63.3 & 70.3 \\
Oxford\_TVG\_CRF\_RNN  \cite{zheng2015conditional} & - & 85.5 & 36.7 & 77.2 & 62.9 & 66.7 & 85.9 & 78.1 & 82.5 & 30.1 & 74.8 & 59.2 & 77.3 & 75.0 & 82.8 & 79.7 & {\bf 59.8} & 78.3 & 50.0 & 76.9 & 65.7 & 70.4 \\
BoxSup-semi-coco \cite{dai2015boxsup} & - & 86.4 & 35.5 & 79.7 & 65.2 & 65.2 & 84.3 & 78.5
& 83.7 & 30.5 & 76.2 & 62.6 & 79.3 & 76.1 & 82.1 & 81.3 & 57.0 & 78.2 & 55.0 & 72.5 & 68.1 & 71.0 \\
DeepLab-MSc-CRF-LargeFOV \cite{chen2014semantic} & 93.1 & 84.4 & 54.5 & 81.5 & 63.6 & 65.9 & 85.1 & 79.1 & 83.4 & 30.7 & 74.1 & 59.8 & 79.0 & 76.1 & 83.2 & 80.8 & 59.7 & 82.2 & 50.4 & 73.1 & 63.7 & 71.6 \\ 
Oxford\_TVG\_CRF\_RNN\_COCO \cite{zheng2015conditional} & - & {\bf 90.4} & {\bf 55.3} & {\bf 88.7} & {\bf 68.4} & 69.8 & {\bf 88.3} & 82.4 & 85.1 & {\bf 32.6} & 78.5 & {\bf 64.4} & 79.6 & {\bf 81.9} & {\bf 86.4} & 81.8 & 58.6 & 82.4 & 53.5 & 77.4 & {\bf 70.1} & {\bf 74.7} \\
\hline \hline
\href{http://host.robots.ox.ac.uk:8080/anonymous/DS0SNS.html}{Cross-Pretrain (Strong)} & 93.4 & 89.1 & 38.3 & 88.1 & 63.3 & 69.7 & 87.1 & {\bf 83.1} & 85.0 & 29.3 & 76.5 & 56.5 & 79.8 & 77.9 & 85.8 & 82.4 & 57.4 & 84.3 & 54.9 & {\bf 80.5} & 64.1 & 72.7 \\ 
\href{http://host.robots.ox.ac.uk:8080/anonymous/HQEH1M.html}{Cross-Joint (Strong)}    & 93.3 & 88.5 & 35.9 & 88.5 & 62.3 & 68.0 & 87.0 & 81.0 & {\bf 86.8} & 32.2 & {\bf 80.8} & 60.4 & {\bf 81.1} & 81.1 & 83.5 & 81.7 & 55.1 & 84.6 & 57.2 & 75.7 & 67.2 & 73.0 \\
\href{http://host.robots.ox.ac.uk:8080/anonymous/7GUG0U.html}{Cross-Pretrain (Strong, Multi-scale net)} & {\bf 93.8} & 88.7 & 53.1 & 87.7 & 64.4 & 69.5 & 85.9 & 81.6& 85.3 & 31.0 & 76.4 & 62.0 & 79.8 & 77.3 & 84.6 & {\bf 83.2} & 59.1 & {\bf 85.5} & 55.9 & 76.5 & 64.3 & 73.6 \\ 
\href{http://host.robots.ox.ac.uk:8080/anonymous/VR0306.html}{Cross-Joint (Strong, Multi-scale net)} & 93.7 & 89.2 & 46.7 & 88.5 & 63.5 & 68.4 & 87.0 & 81.2 & 86.3 & {\bf 32.6} & 80.7 & 62.4 & 81.0 & 81.3 & 84.3 & 82.1 & 56.2 & 84.6 & {\bf 58.3} & 76.2 & 67.2 & 73.9 \\
\hline
 \end{tabular}
} 
 \caption{VOC 2012 \textsl{test} performance using strong PASCAL and strong MS-COCO annotations (Sec.~4.5 of main paper). Links to the PASCAL evaluation server are included in the PDF.}
 \label{tab:sup_cross_data_test}
\end{table*}

{\small
\bibliographystyle{ieee}
\bibliography{egbib}
}

\end{document}